\documentclass{article} 
\usepackage{iclr2026_conference,times}

\usepackage{amsmath,amsfonts,bm}

\def\eqref#1{equation~\ref{#1}}

\def\1{\bm{1}}

\DeclareMathAlphabet{\mathsfit}{\encodingdefault}{\sfdefault}{m}{sl}
\SetMathAlphabet{\mathsfit}{bold}{\encodingdefault}{\sfdefault}{bx}{n}

\usepackage{hyperref}
\usepackage{wrapfig}
\usepackage{times}  
\usepackage{helvet}  
\usepackage{courier}  
\usepackage{graphicx} 
\usepackage{natbib}  
\usepackage{caption} 
\usepackage{subcaption} 
\usepackage{amsmath}
\usepackage{amsthm}
\usepackage{algorithm}
\usepackage{algorithmic}
\usepackage{array,booktabs}
\usepackage[table]{xcolor} 
\usepackage{amssymb}
\usepackage{textcomp}
\usepackage{multirow}
\usepackage{makecell}
\usepackage{enumitem}
\usepackage{xcolor} 
\usepackage{pifont} 
\usepackage[switch]{lineno}

\newtheorem{theorem}{Theorem}
\newtheorem{lemma}[theorem]{Lemma}
\newtheorem{corollary}[theorem]{Corollary}
\newtheorem{definition}[theorem]{Definition}

\definecolor{darkgreen}{rgb}{0.0, 0.39, 0.0}
\definecolor{lightgreen}{rgb}{0.5, 1.0, 0.5}

\definecolor{lightgray}{rgb}{0.92,0.92,0.92}
\usepackage{newfloat}
\usepackage{listings}
\DeclareCaptionStyle{ruled}{labelfont=normalfont,labelsep=colon,strut=off} 
\floatstyle{ruled}
\newfloat{listing}{tb}{lst}{}
\floatname{listing}{Listing}

\title{Flatness Guided Test-Time Adaptation for Vision-Language Models}

\iclrfinalcopy

\usepackage{marvosym}

\author{Aodi Li$^1$, Liansheng Zhuang$^{1,2}$\textsuperscript{(\Letter)}, Xiao Long$^1$,  Houqiang Li$^{2}$ \& Shafei Wang$^3$  \\
$^1$School of Cyber Science and Technology, University of Science and Technology of China \\
$^2$National Engineering Laboratory for Brain-Inspired Intelligence Technology and 
Applications, \\
\: University of Science and Technology of China \quad 
$^3$Peng Cheng Laboratory, Shenzhen, China\\ 
\texttt{aodili@mail.ustc.edu.cn, lszhuang@ustc.edu.cn} }

\begin{document}

\maketitle

\begin{abstract}

Test-time adaptation (TTA) of Vision-Language Models (VLMs) has emerged as a technique for tackling distribution shifts during the test time. 
Recent research indicates that the test-time adaptation is intrinsically linked to the model's training history.
However, existing TTA methods, such as Test-time Prompt Tuning, often design adaptation strategies in isolation from the models' training characteristics, which degrade their performance. 
This paper argues that the flatness acquired via sharpness-aware training is an efficient clue for the test-time adaptation of VLMs. 
Built on this insight, this paper proposes a novel Flatness-Guided Adaptation framework (FGA) for VLMs to cohesively unify training and test-time procedures. Its core idea is to leverage the alignment between the training minimum and test loss flat regions to guide the adaptation process.
Specifically, our FGA consists of a prompt-tuning stage and a test-time adaptation stage. In the tuning stage, a Sharpness-Aware Prompt Tuning method is utilized to identify the training flat minimum, offering a geometric clue of flatness for subsequent adaptation. In the test stage, a Sharpness-based Test Sample Selection approach is proposed to ensure the alignment of flat minima between the training and each augmented test sample's loss landscape. 
In comparison to existing TTA methods, our FGA avoids the expensive prompt parameter updates during test time, and substantially reduces the computation overhead.
Extensive experiments on both domain generalization and cross-dataset benchmarks demonstrate that our FGA achieves superior performance over prevalent TTA methods. 
Notably, when employing a ViT-B/16 image encoder, FGA even outperforms TPT+CoOp by an average of 4.88\% across all four ImageNet out-of-domain variants.
\end{abstract}
\section{Introduction}

Recent advancements in vision-language pretraining, such as CLIP~\citep{CLIP}, have generated new opportunities for developing foundational models in vision tasks~\citep{VLone,VLtwo}. These models, trained on extensive collections of image-text pairs, can learn and represent a diverse range of visual concepts. By means of well-designed prompts, they can be applied to downstream tasks in a zero-shot manner without requiring task-specific data~\citep{GLIP,Hier,styleclip}. 
Consequently, various prompt tuning methods~\citep{coop,cocoop} are proposed to directly learn prompts using training data from downstream tasks. Though these methods find better prompts compared to hand-crafted ones, the learned prompts are limited to the training distribution and may have limited generalization beyond that. 

\setlength{\textfloatsep}{0.3cm} 
\begin{figure}[t]
    \centering
    \vspace{-0.5cm}
    \begin{subfigure}[b]{0.4\textwidth} 
        \centering
        \includegraphics[width=\textwidth]{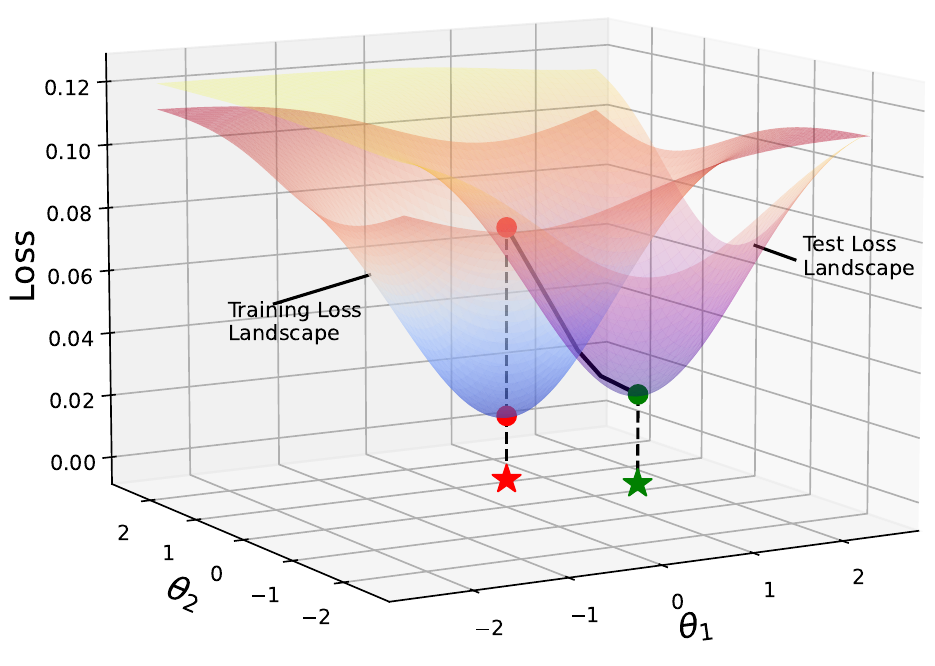}
        \caption{\textbf{Traditional TTA Methods}}  
        \label{fig:landscape_train}
    \end{subfigure}
    \hspace{0.2cm}
    \begin{subfigure}[b]{0.4\textwidth} 
        \centering
        \includegraphics[width=\textwidth]{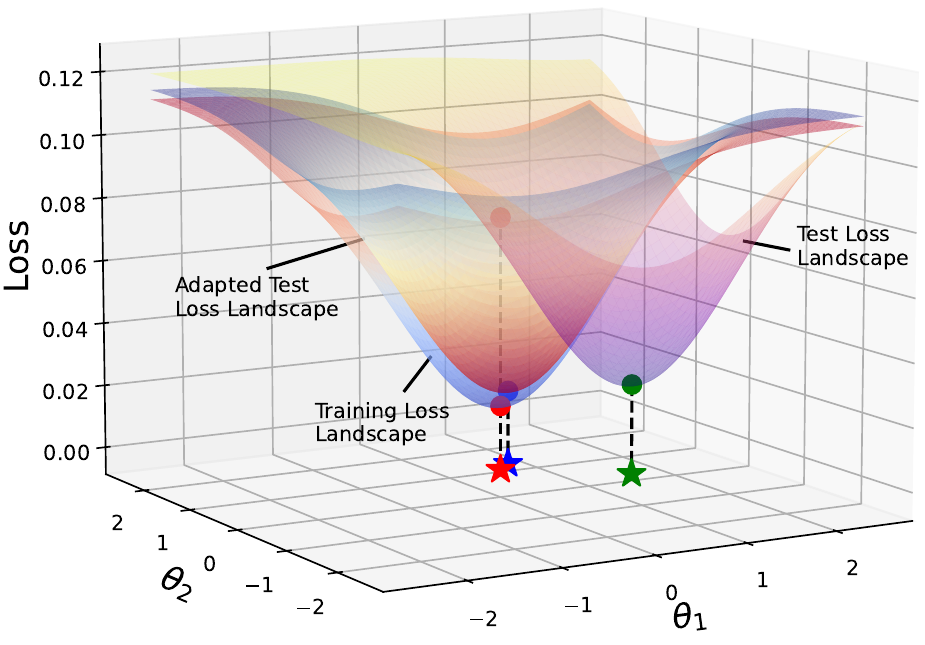}
        \caption{\textbf{Our FGA}}
        \label{fig:landscape_test}
    \end{subfigure}
    \caption{\textbf{Comparison of conventional TTA methods and Flatness-Guided Adaptation (FGA).} 
    (a) Traditional TTA methods treat the test landscape as static, aiming to optimize parameters to achieve the test flat minimum 
    (\textcolor{darkgreen}{\scalebox{0.7}{\ding{72}}}), using the training minimum (\textcolor{red}{\scalebox{0.7}{\ding{72}}}) as an initialization. 
    (b) Our FGA keeps parameters unchanged during testing. Instead, it adjusts the test landscape to a position where the training minimum 
    (\textcolor{red}{\scalebox{0.7}{\ding{72}}})
    is already very close to the minimum (\textcolor{blue}{\scalebox{0.7}{\ding{72}}}) of the adapted test landscape.}
    \label{fig:tta_12}
\end{figure}

To address this issue, several studies~\citep{ContrastiveTTA,PFTTA,tent} have attempted to develop test-time adaptation (TTA) methods, which aim to rapidly adjust pre-trained models to unlabeled test data streams during inference. Among various TTA strategies, methods based on Test-Time Prompt Tuning (TPT)~\citep{TPT}, which optimize a set of learnable prompts via entropy minimization on augmented test views, have demonstrated promising performance and gained significant attraction ~\citep{TPT,DiffTPT,C-TPT}. 
Recent research indicates that the test-time adaptation is intrinsically influenced by the model’s training history~\citep{conjy}.
However, most existing TTA methods, including TPT-based methods, design adaptation strategies in isolation, treating the test phase as a standalone optimization problem disconnected from the model's training history. 
This isolation from the training phase may fail to exploit valuable geometric and representational properties inherent in the pre-trained model, leading to suboptimal test-time adaptation.

To improve model generalization, seeking flat minima within the training loss landscape has emerged as an effective training strategy over the past few years~\citep{SAM,ASAM,FisherSAM}. 
It is widely observed that parameters residing in flat minima tend to generalize better to out-of-distribution data~\citep{SWAD,DISAM,SFT,SAMDG2024} than those sharp ones. 
Nonetheless, conventional TTA methods often ignore the influence of sharpness-aware training on the test-time adaptation. While Sharpness-Aware Minimization (SAM)~\citep{SAM} seeks flat regions during training, its principle is rarely extended to guide test-time adaptation in a unified framework. This disconnection leads to computationally expensive test-time optimizations (e.g., backpropagation in TPT~\citep{TPT}) that are agnostic to loss geometric structure and often yield suboptimal generalization.
This paper argues that flatness is not merely a desirable property during training but a powerful clue that can dictate test-time adaptation.

Inspired by this insight, this paper proposes a novel \textit{Flatness-Guided Adaptation} (FGA) framework, which cohesively unifies training and test-time procedures from the perspective of loss landscape geometry. 
It mainly leverages the alignment between the training minimum and test loss flat regions to guide the adaptation process (see Figure~\ref{fig:tta_12}).
Specifically, our FGA framework consists of two synergistic stages: 
(1) In the prompt tuning stage, a \textit{Sharpness-Aware Prompt Tuning} (SAPT) method is utilized to fine-tune the prompts on the downstream training dataset, aiming at seeking the training flat minimum. Since flatter minima generally indicate better model generalization than sharper ones~\citep{SharpGenelink,SharpGenelinkTwo,SharpGenelinkThree,SAM}, 
the minimization of sharpness not only improves model generalization but also provides a test-time criterion to measure the alignment of flat minima within the training and test loss landscapes. 
(2) In the test-time stage, FGA leverages the geometric clue of flatness acquired via SAPT. For a given test sample, a \textit{Sharpness-based Test Sample Selection} (STSS) method is proposed to intelligently select its augmented views based on the sharpness score of their loss landscapes around the training flat minimum. This ensures that the final prediction is derived from a test-time loss landscape whose flat minima align with those identified during training.
During this process, loss landscapes are efficiently altered through data augmentations.
In comparison with existing TTA methods, our FGA avoids the expensive prompt parameter updates during test time,
eliminating the computational overhead of adaptation and offering a more plausible adaptation strategy.
Theoretical analysis suggests that using the sharpness-based metric will help distinguish the proximity of test samples to the training distribution. 
The closer an augmented sample is to the training distribution, the smaller its sharpness-based score is likely to be. 
Since models tend to generate more reliable results for data closer to the training distribution, FGA significantly improves the generalization ability of vision-language models. 
Extensive experiments on domain generalization~\citep{oodbenchmark} and cross-dataset~\citep{cocoop} benchmarks demonstrate the superior performance of FGA over prevailing TTA methods.

Our main contributions can be summarized as follows:
\begin{itemize}[leftmargin=0.5cm, itemsep=5pt, topsep=0pt, parsep=0pt]
    \item  A novel Flatness-Guided Adaptation (FGA) framework is proposed to cohesively unify training and test-time procedures for vision-language models. 
    By ensuring the alignment of the model's training flat minimum with flat regions in test loss landscapes, it significantly enhances the generalization capabilities of VLMs under distribution shifts.
    \item Theoretical analysis is presented to offer a clearer insight into how sample selection at test time improves the reliability of predictions.
    \item Extensive experiments on domain generalization and cross-dataset benchmarks demonstrate the superior performance of FGA over other prevalent TTA methods, while significantly eliminating the computational overhead.
\end{itemize}

\section{Related Work}

\textbf{Test-time adaptation (TTA) of vision-language models.}
Vision-language models like CLIP have shown strong performance in various tasks. To enhance CLIP’s transfer learning for downstream classification tasks, methods like text prompt learners (e.g., CoOp~\citep{coop} and CoCoOp~\citep{cocoop}) and visual adapters (e.g., Tip-Adapter~\citep{tip}) have been proposed. 
However, these methods struggle with distribution misalignment between pre-training and test data.
Test-time adaptation (TTA) methods address this by adjusting models during testing, with two main streams~\citep{align}: the first modifies the training process using a self-supervised proxy task, such as image rotation prediction, and uses it to guide test-time optimization (e.g., Test-Time Training~\citep{TTT} and TTT++~\citep{TTTPlus}); the second adapts models without altering the training process (e.g., TPT~\citep{TPT}, which uses entropy minimization to learn adaptive parameters during testing).  DiffTPT~\citep{DiffTPT} introduces a diffusion model to generate diverse augmentations for further improvements. PromptAlign~\citep{align} adds an explicit term to align the learned distributions with that of test data. Meanwhile, online methods like TDA~\citep{TDA} and DPE~\citep{DPE} use a key-value cache or prototype set to adapt progressively to test data. 
They benefit from information aggregated during testing but are unsuitable for single test-sample scenarios, unlike TPT-based methods.
This paper proposes a novel Flatness-Guided Adaptation (FGA) framework that leverages the geometry of loss landscapes to enhance CLIP's generalization and inference efficiency in single test-sample adaptation scenarios. By avoiding backpropagation and parameter updates during testing, FGA significantly reduces computational overhead while achieving robust out-of-domain performance.

\textbf{Generalization from a loss landscape view.} 
In recent years, optimization techniques aimed at flat minima in loss landscapes have surged to improve the generalization of deep models~\citep{SharpGenelink,SharpGenelinkTwo,SharpGenelinkThree}. Among them, SAM~\citep{SAM}, which focuses on finding parameters located in regions of the loss landscape with consistently low loss values, has gained significant attention for its effectiveness and scalability. 
To seek flatter minima, numerous SAM variants, such as ASAM~\citep{ASAM} and FisherSAM~\citep{FisherSAM}, have already been developed over the past few years.
This concept of flat minima has also been extended to improve the out-of-domain generalization of deep models~\citep{SAMDG2024,SWAD,SFT}. However, most of them focus on the training stage.
SAR~\citep{SAR} and SoTTA~\citep{SoTTA}, two online test-time adaptation (TTA) methods, both utilize sharpness-aware minimization at test time to improve robustness by seeking flat minima. Yet, they operate solely during testing without accounting for training-testing sharpness interactions.
In contrast, our FGA applies sharpness-aware minimization during training to establish flatness as a criterion for subsequent alignment, then at test time adapts by adjusting loss landscapes through augmentation selection (without updating model parameters), and preserving the pre-trained flat minimum’s optimality on adapted test loss landscapes.

\section{Methodology}
\subsection{Preliminaries}
\textbf{Contrastive Language-Image Pre-training.}
CLIP~\citep{CLIP} primarily comprises a Text Encoder $\boldsymbol{E}_t$ and an Image Encoder $\boldsymbol{E}_v$. The Image Encoder is available in two architectures:
\begin{wrapfigure}{r}{0.5\textwidth}
    \centering
    \includegraphics[width=\linewidth]{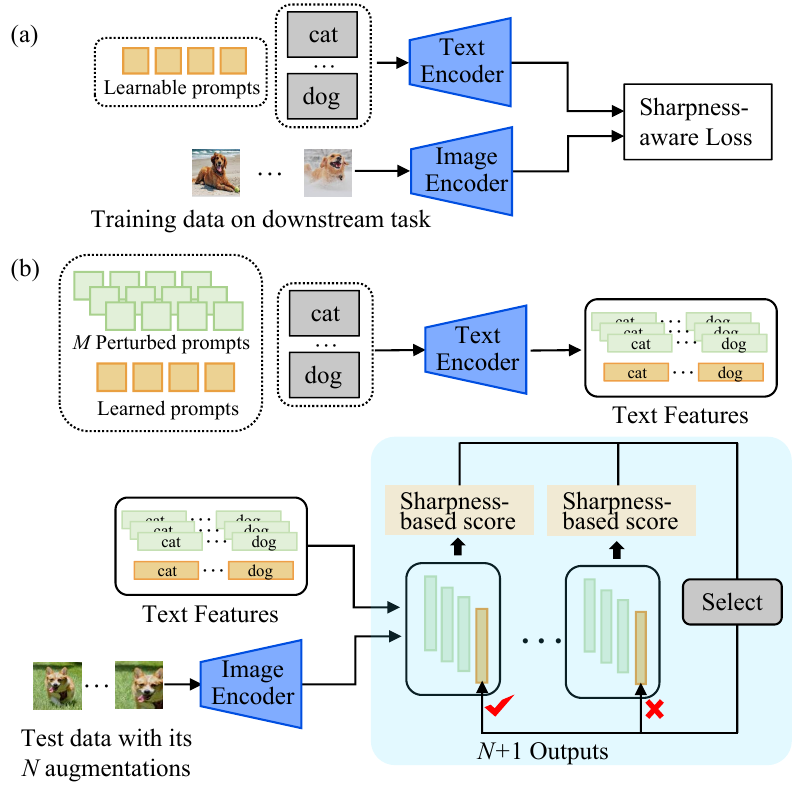}
    \vspace{-0.3cm}
    \caption{\textbf{Overview of our Flatness-Guided Adaptation (FGA).} It consists of two synergistic mechanisms:
        (a) Sharpness-aware Prompt Tuning: It optimizes the model parameters to reduce the loss value and sharpness, enabling stable and effective adaptations during test time without direct access to training data.
        (b) Sharpness-based Test Sample Selection: It introduces a selection mechanism to identify augmented test samples that ensure the training flat minimum aligns with those in their loss landscapes, enabling more confident predictions.
    } 
    \label{fig:overview}
\end{wrapfigure}
one based on ResNet~\citep{resnet} and the other using the popular Vision Transformer (ViT)~\citep{vitbase}. This encoder transforms an input image $\boldsymbol{x}$ into its feature representation, i.e., $\boldsymbol{e}_i = \boldsymbol{E}_v (\boldsymbol{x})$.
For a classification task with $K$ classes, the $k$-th class label is formatted into the textual prompt ``a photo of a [$k$-th class]'', which is then tokenized into a sequence $\boldsymbol{l}_k = (t_{\mathrm{SOS}}, t_1, t_2, \ldots, t_L, c_k, t_{\mathrm{EOS}})$.
Here, $t_{\mathrm{SOS}}$ and $t_{\mathrm{EOS}}$ represent the embeddings of the start and end tokens, while $t_1, t_2, \ldots, t_L$ corresponds to the phrase ``a photo of a'', and the token $c_k$ denotes the specific description of the $k$-th class. The text encoder of CLIP, designed as a Transformer architecture, processes these tokens to generate text features: $\boldsymbol{e}_{t,k} = \boldsymbol{E}_t (\boldsymbol{l}_k)$.
During the pre-training stage, CLIP is trained on the WIT dataset~\citep{CLIP} through a contrastive learning approach. 
In this setup, each image is paired with its corresponding text sentence as a positive sample, while all other image-text combinations are treated as negative samples. The goal of the contrastive learning objective is to enhance the cosine similarity of positive pairs while reducing that of negative pairs.
In the classification stage, all classes in the dataset are converted to text, and the cosine similarity between image embeddings and text embeddings is computed to determine the probability of an image belonging to each category:
\begin{align}
\mathbb{P}\left(y_k \mid \boldsymbol{x}\right)=\frac{\exp \left(\operatorname{sim}\left(\boldsymbol{e}_{t,k} \cdot \boldsymbol{e}_{i}\right) \tau\right)}{\sum_{j=1}^K \exp \left(\operatorname{sim}\left(\boldsymbol{e}_{t,j} \cdot \boldsymbol{e}_i\right) \tau\right)}, 
\end{align}
where $\tau$ is the temperature of the softmax.

\textbf{Prompt tuning.} 
Prompt tuning has emerged as a popular tuning method for Transformer-based models in downstream tasks. This approach does not modify the model parameters; rather, it changes the input to the model, making it highly efficient. Specifically, instead of using the template ``a photo of a [class]'', it replaces the tokens associated with the hand-crafted prompts (``a photo of a'') with learnable parameters $\boldsymbol{p}=(p_1,\ldots,p_L)$, which are then updated based on the dataset used for downstream tasks.

\textbf{Test-time prompt tuning.} 
Prompts optimized on downstream training data are prone to overfitting and often generalize poorly under distribution shifts. To mitigate this, Test-Time Prompt Tuning (TPT)~\citep{TPT} fine-tunes a specific prompt for each individual test sample.
During testing, multiple augmented views of the test samples are generated. Then, predictions with entropy below a predetermined threshold are kept, while others are discarded using a confidence filter. The averaged entropy of selected predictions is then used as a loss function to update the prompts.

\subsection{Flatness-Guided Adaptation}

Our proposed Flatness-Guided Adaptation (FGA) framework fundamentally offers a unified, loss landscape-centric methodology that seamlessly bridges the training and test phases.
This approach leverages the flatness as a universal guiding principle to enhance both generalization during training and robust adaptation during inference under distribution shifts.
As illustrated in Figure \ref{fig:overview}, FGA integrates two complementary mechanisms that operate in concert: sharpness-aware prompt tuning (SAPT) during training and sharpness-guided test sample selection (STSS) during inference. 
This section will elaborate on the key idea and technical details of the framework.

\subsubsection{Sharpness-Aware Prompt Tuning} 

FGA mainly exploits the alignment between training and test flat minima for the efficient adaptation of VLMs.
However, a major challenge in achieving this alignment arises from inherent data constraints. Test samples remain unknown during training, and at test time, training data becomes inaccessible due to storage limitations and privacy requirements.
To overcome this, FGA focuses on the intrinsic properties of the training minimum, with the goal that the test loss landscape exhibits these same desirable characteristics:
(1) The training minimum should correspond to a low loss value, indicating that the model is effectively learning from the data; (2) The training minimum may display implicit biases, such as reduced sharpness, which are often beneficial for improved generalization.

Traditional training methods for optimizing the prompts, such as CoOp~\citep{coop}, typically use cross-entropy loss to fine-tune the prompt $\boldsymbol{p}$:
\begin{equation}
    \ell_{\mathrm{CE}}(\boldsymbol{p})  = -\sum_{i=1}^{n} \log \mathbb{P}_{\boldsymbol{p}}(y_i|\boldsymbol{x}_i),  
\end{equation}
where 
$p_{\boldsymbol{p}}(y_i|\boldsymbol{x}_i)$ represents the predictive probability that $\boldsymbol{x}_i$ belongs to the its true label class. 
While standard SGD tends to find flat minima, methods such as SAM~\citep{SAM} can enhance this implicit bias through explicit perturbation-aware optimization.
To enable more precise alignment during testing, we adopt Sharpness-aware Prompt Tuning (SAPT) during training, which jointly minimizes both the loss and its ``sharpness'':
\begin{equation}
    \ell_{\mathrm{SAPT}}(\boldsymbol{p}) =  \ell_{\mathrm{CE}}(\boldsymbol{p})+ \lambda \max_{|| \boldsymbol{\epsilon} || \leq \rho} 
    \left[
    \ell_{\mathrm{CE}}(\boldsymbol{p} + \boldsymbol{\epsilon}) - \ell_{\mathrm{CE}}(\boldsymbol{p})
    \right].
\end{equation}
The first and second terms above represent the loss value and loss sharpness, with $\lambda$ acting as a hyperparameter to balance them.
Similar to previous studies~\citep{SAM,ASAM}, sharpness is defined as the sensitivity of the training loss to small perturbations $\boldsymbol{\epsilon} $ (with a norm less than $ \rho $) added to the prompts $ \boldsymbol{p} $. 
Since the perturbation strength $\rho$ is small enough, we can apply a Taylor expansion to approximately solve for the optimal perturbation $ \boldsymbol{\epsilon}^\star $:
\begin{align}
    \boldsymbol{\epsilon}^\star 
&= 
\mathop{\mathrm{arg\,max}}_{\|\boldsymbol{\epsilon}\| \leq \rho} 
    \ell_{\mathrm{CE}}(\boldsymbol{p} + \boldsymbol{\epsilon}) - \ell_{\mathrm{CE}}(\boldsymbol{p}) \approx 
\mathop{\mathrm{arg\,max}}_{\|\boldsymbol{\epsilon}\| \leq \rho} 
    \boldsymbol{\epsilon}^\top \nabla_{\boldsymbol{p}}\ell_{\mathrm{CE}}(\boldsymbol{p}) 
= \rho \frac{\nabla_{\boldsymbol{p}}\ell_{\mathrm{CE}}(\boldsymbol{p})}{\|\nabla_{\boldsymbol{p}}\ell_{\mathrm{CE}}(\boldsymbol{p})\|}.
\end{align}
During training via stochastic gradient descent, the contribution from $ \nabla_{\boldsymbol{p}}\boldsymbol{\epsilon}^\star $ can be disregarded due to the minor perturbation strength $ \rho $. 

In this way, SAPT not only yields robust prompts that enhance generalization but also provides the sharpness measure as additional information for adaptation during testing.

\subsubsection{Sharpness-based Test Sample Selection}
Through sharpness-aware prompt tuning, the prompts are positioned at a flat minimum within the training loss landscape. 
To avoid computationally expensive gradient descent during inference, we keep the pre-trained prompt fixed and instead adapt the test loss landscapes such that the well-trained prompt from the downstream training dataset (i.e., the training flat minimum) coincides with the flat minimum in the adapted test landscape, as illustrated in Figure \ref{fig:tta_12}. 

To achieve this alignment, we propose a Sharpness-based Test Sample Selection (STSS) method. 
STSS utilizes data augmentations to create multiple test loss landscapes for each sample. 
By selecting augmented samples that align the training minimum with flat minima in their respective loss landscapes, we ensure the training minimum remain optimal. Given that such alignment typically corresponds to small loss values and reduced loss sharpness in these test landscapes, STSS introduces a sharpness-based score as a metric.
To mitigate the computational burden of backpropagation in calculating sharpness, we redefine it as the maximum variation in the loss resulting from $R$ random perturbations:
\begin{align}
    &\ell_{\mathrm{STSS}}(\boldsymbol{p}) = 
    \ell_{\mathrm{SRG}}(\boldsymbol{p}) +\lambda\max_{r=1,\ldots,R;\boldsymbol{\epsilon_r}\sim\mathcal{N}} 
    \left[\ell_{\mathrm{SRG}}\left(\boldsymbol{p} + \rho'\frac{\boldsymbol{\epsilon_r}}{\|\boldsymbol{\epsilon_r}\|}\right) - \ell_{\mathrm{SRG}}(\boldsymbol{p})\right].
\end{align}
Here, $\ell_{\mathrm{STSS}}$ represents the sharpness-based score used to select the most reliable augmented test samples, and  $\ell_{\mathrm{SRG}}$ denotes a surrogate loss function when test labels are unavailable, such as entropy~\citep{tent,conjy}.  
The perturbation direction is expressed by $\boldsymbol{\epsilon_r}/\|\boldsymbol{\epsilon_r}\|$, where $\boldsymbol{\epsilon_r}$ is drawn from the standard normal distribution $\mathcal{N}$. The quantity $\rho'$ controls the magnitude of perturbations during testing. 
To obtain $\ell_{\mathrm{SRG}}\left(\boldsymbol{p} + \rho'\frac{\boldsymbol{\epsilon_r}}{\|\boldsymbol{\epsilon_r}\|}\right)$, we first obtain text features for each category ($\boldsymbol{e}_{t,k,r}$) through the forward pass  of the text encoder:
\begin{equation}    [\boldsymbol{e}_{t,k,1},\ldots,\boldsymbol{e}_{t,k,R}] = \boldsymbol{E}_t( [\boldsymbol{l}_{k,1}, \ldots,  \boldsymbol{l}_{k,R}]),
\end{equation}
where  
$\boldsymbol{l}_{k,r}$ denotes the input token related to the $k$-th category with the $r$-th pertuation. 
Notably, the additional computational cost of this step is minimal, as text features only need to be computed once per test category. Then, the surrogate loss for perturbed prompts is:
\begin{align}
\ell_{\mathrm{SRG}}\left(\boldsymbol{p} + \rho'\frac{\boldsymbol{\epsilon_r}}{\|\boldsymbol{\epsilon_r}\|}\right)=
-\sum_{k=1}^K \mathbb{P}_r(y_{k}|\boldsymbol{x}) \log \mathbb{P}_r(y_{k}|\boldsymbol{x}),
\end{align}
with probabilities derived from cosine similarity:
\begin{align}
\mathbb{P}_r(y_{k}|\boldsymbol{x}) =
\frac{\exp \left(\operatorname{sim}\left(\boldsymbol{e}_{t,k,r} \cdot \boldsymbol{e}_{i}\right) \tau\right)}{\sum_{j=1}^K \exp \left(\operatorname{sim}\left(\boldsymbol{e}_{t,j,r} \cdot \boldsymbol{e}_i\right) \tau\right)}.  
\end{align}
Finally, the final prediction aggregates votes from the top $s$ augmented samples with the lowest sharpness-based scores, which are more reliable predictions according to the theoretical analysis in the next section.

\section{Theoretical Analysis}
This section provides a theoretical explanation of how our method improves test-time classification.
Let's begin with the following problem: During training, the model learns from data sampled independently and identically from distribution $\mathcal{S}$; During testing, however, data is drawn from two distinct distributions $\mathcal{T}_1$ and $\mathcal{T}_2$. Then, the question is: \textit{How can we distinguish between these test distributions and determine on which one the model will perform more reliably?
}

To address this, we first derive an upper bound for the generalization error, which quantifies the model's performance on unseen data from $\mathcal{T}_1$ and $\mathcal{T}_2$. We will then explore how, when the test distributions are sufficiently distinguishable, FGA can effectively distinguish between them. This is crucial because, as we will show, when the test distribution closely resembles the training distribution, the generalization error bound decreases, leading to more accurate predictions.

\setlength{\tabcolsep}{3mm}
\begin{table*}[t]
\centering
\caption{\textbf{Results on datasets with natural distribution shifts.} We report
top-1 accuracy (\%) for each method across five datasets, using the CLIP-ViT-B/16 backbone. We highlight the best results in \textbf{bold} and \underline{underline} the second-best results. The abbreviation ``IN'' means the ImageNet dataset. ``$\dagger$'' denotes results reproduced by adapting the method to the single-sample setting.
}
\small
\scalebox{0.85}{\begin{tabular}{l|ccccc|c|c}
\toprule
Algorithm & IN & IN-A & IN-V2 & IN-R & IN-Sketch & Avg. & OOD Avg. \\
\midrule
CLIP-ViT-B/16~\citep{CLIP} & 68.34 & 49.89 & 61.88 & 77.65 & 48.24 & 61.20 & 59.42 \\
\midrule
CoOp~\citep{coop} & 71.51 & 49.71 & 64.20 & 75.21 & 47.99 & 61.72 & 59.28 \\

CoCoOp~\citep{cocoop} & 71.02 & 50.63 & 64.07 & 76.18 & 48.75 & 62.13 & 59.91\\

Tip-Adapter~\citep{tip} & 70.75 & 51.04 & 63.41 & 77.76 & 48.88 & 62.37 & 60.27\\
\midrule
TPT~\citep{TPT} & 69.70 & 53.67 & 64.30 & 73.90 & 46.40 & 61.59 & 59.57 \\
DiffTPT~\citep{DiffTPT}  & 70.30 & 55.68 & 65.10 & 75.00 & 46.80 & 62.58 & 60.64 \\
C-TPT~\citep{C-TPT} & - & 52.90 & 63.40 & 78.00 & 48.50 & - & 60.70  \\
ZERO~\citep{ZERO} & 69.06 & {61.35} & 64.13 & 77.28 & 48.29 & 64.02 & 62.76 \\
MTA~\citep{MTA} & 69.29 & 57.41 & 63.61 & 76.92 & 48.58 & 63.16 & 61.63\\
PromptAlign$^\star$~\citep{align} & - & 59.37 & 65.29 & {79.33} & {50.23} & - & {63.56} \\
TDA~\citep{TDA} & 69.51 & 60.11 & 64.67 & {80.24} & {50.54} & {65.01} & {63.89} \\
DPE~\citep{DPE} & 71.91  & 59.63 & 65.44 & \underline{80.40} & \textbf{52.26} & {65.93} & {64.43}\\
TPT~\citep{TPT}+CoOp  & {73.30} & 56.88 & {66.60} & 73.80 & 49.40 & 64.00 & 61.67 \\
DiffTPT~\citep{DiffTPT}+CoOp  & \textbf{75.00} & 58.09 & 66.80 & 73.90 & 49.50 & {64.66} & 62.07 \\
C-TPT~\citep{C-TPT}+CoOp & 72.90 & 52.73 & 65.61 & 76.46 & 48.63 & 63.27 & 60.86 \\
ZERO~\citep{ZERO}+CoOp & 73.61 & {63.17} & 66.82 & 77.71 & 48.52 & 65.97 & 64.06 \\
MTA~\citep{MTA}+CoOp  & 73.99 & 59.29  & 66.97 &  78.20 &  49.96 &  65.68 & 63.61 \\
SAR$^\dagger$~\citep{SAR}+CoOp & 73.03 & 55.35 & 65.89 & 77.09 & 48.65 & 64.00 & 61.75  \\
\midrule
\textbf{SAPT+CoOp} & 70.79 & 51.04 & 64.41 & 77.66 & 49.31 & 62.64 & 60.61 \\
\textbf{STSS+CoOp} & 73.99 & \underline{64.00} & \underline{67.11} & 77.92 & 49.36 & \underline{66.48} & \underline{64.60} \\
\textbf{FGA (Ours)} & \underline{74.01} & \textbf{65.90} & \textbf{67.23} & \textbf{81.24} & \underline{51.81} & \textbf{68.04} & \textbf{66.55} \\
\bottomrule
\end{tabular}}
\label{tab:DG}
\end{table*}

\begin{theorem}[Generalization Bound]\label{th:GB}
    Consider a vector-valued function class $\mathcal{F}_v = \{\boldsymbol{f}_{\boldsymbol{\theta}}(\cdot):\mathcal{X}\to\mathbb{R}^K\}$ and a bounded loss function $\ell: \mathbb{R}^{K} \times \mathcal{Y} \to [0, M]$. 
    Define the worst-case loss $\ell^{\rho}$ within a $\rho$-radius perturbation as
    \begin{equation}
        \ell^{\rho}(\boldsymbol{f}_{\boldsymbol{\theta}}(\boldsymbol{x}), y) = \max_{\|\boldsymbol{\epsilon}\|_2 \leq \rho} \, \ell(\boldsymbol{f}_{\boldsymbol{\theta} + \boldsymbol{\epsilon}}(\boldsymbol{x}), y).
    \end{equation}
    Assume $\ell^{\rho}$ is $\mu$-Lipschitz continuous with respect to $\boldsymbol{f}$.
    Let $\mathcal{S}$ and $\mathcal{T}$ denote the training and test distributions, respectively. Denote $\mathcal{S}_n=\{\boldsymbol{x}_i,y_i\}_{i=1}^{n}$ as $n$ i.i.d. training samples drawn from distribution $\mathcal{S}$.  Then, with probability at least $1 - \delta$, the following generalization bound holds:
    \begin{align}
        \mathbb{E}_{\mathcal{T}}\bigl[ \ell^{\rho}(\boldsymbol{f}_{\boldsymbol{\theta}}(\boldsymbol{x}_{\scriptscriptstyle{\mathcal{T}}}), y_{\scriptscriptstyle{\mathcal{T}}}) \bigr] 
        &\leq \frac{M}{2}\, d\left(\mathcal{S}; \mathcal{T}\right) 
        + \hat{\ell}^{\rho}_{\mathcal{S}_n}(\boldsymbol{f}_{\boldsymbol{\theta}}) 
        + 2\sqrt{2}\,\mu\, R_n(\mathcal{F}, \mathcal{S}) 
        + M \sqrt{\frac{\log(1/\delta)}{2n}},
        \label{eq:upbound}
    \end{align}
    where $(\boldsymbol{x}_{\scriptscriptstyle{\mathcal{T}}}, y_{\scriptscriptstyle{\mathcal{T}}})$ is a random pair following the distribution $\mathcal{T}$; $\hat{\ell}^{\rho}_{\mathcal{S}_n}(\boldsymbol{f}_{\boldsymbol{\theta}}) := \frac{1}{n}\sum_{i=1}^n \ell^{\rho}\bigl(\boldsymbol{f}_{\boldsymbol{\theta}}(\boldsymbol{x}_i), y_i\bigr)$ denotes the empirical loss over $\mathcal{S}_n$; $d(\mathcal{S}; \mathcal{T})$ is the distribution discrepancy measure defined in Appendix~A; and $R_n(\mathcal{F}_v, \mathcal{S})$ represents the Rademacher complexity of $\mathcal{F}_v$ on $\mathcal{S}$.
\end{theorem}

In the following, we will show that when the two test distributions are sufficiently distinguishable—compared with the tightness of the above upper bound—we can effectively differentiate between them. 
To proceed with this analysis, we first introduce the concepts of bound tightness and distribution separability.

\begin{definition}[$\beta$-tightness]
    Given a random variable $X$ and a confidence level $\delta \in (0,1)$, let $\alpha$ be an upper bound such that $\mathbb{P}(X \leq \alpha) \geq 1 - \delta$.
    If there exists an oracle upper bound $\alpha^\star$ satisfying $\mathbb{P}(X \leq \alpha^\star) = 1 - \delta$, then $\alpha$ is said to be $\beta$-tight with $\beta = |\alpha - \alpha^\star|$.
\end{definition}

\begin{definition}[$\gamma$-separability]
    Let $\mathcal{T}_1$ and $\mathcal{T}_2$ be two test distributions and $\mathcal{S}$ the training distribution. 
    Denote by $d(\cdot; \mathcal{S})$ the discrepancy between a distribution and $\mathcal{S}$.
    Then $\mathcal{T}_1$ and $\mathcal{T}_2$ are called $\gamma$-separable if
    $\bigl| d(\mathcal{T}_1; \mathcal{S}) - d(\mathcal{T}_2; \mathcal{S}) \bigr| > \gamma.$
\end{definition}

\begin{theorem}\label{th:main}
Consider a vector-valued function class $\mathcal{F}_v = \{\boldsymbol{f}_{\boldsymbol{\theta}}(\cdot):\mathcal{X}\to\mathbb{R}^K\}$, where the parameters \(\boldsymbol{\theta}\) lie in a set ${\Theta}$ such that the loss function is bounded within \([0, M]\).  
Let \(\boldsymbol{f} = (f_1, \dots, f_K)\) and \(\mathbf{y} = (\mathrm{y}_1, \dots, \mathrm{y}_K)\) be probability distributions over a finite set \(\{1, \dots, K\}\), with \(f_i, \mathrm{y}_i \geq \eta > 0\) for all \(i\). Denote by \(H(\boldsymbol{f})\) the entropy and by \(H(\mathbf{y}, \boldsymbol{f})\) the cross entropy.  Assume the worst-case cross entropy $H^{\rho}(\mathbf{y},\boldsymbol{f})$ is $\mu$-Lipschitz continuous with respect to $\boldsymbol{f}$ over ${\Theta}$.
Given a training distribution \(\mathcal{S}\) and two \(\gamma\)-separable test distributions \(\mathcal{T}_1\) and \(\mathcal{T}_2\), assume \(d(\mathcal{S}, \mathcal{T}_1) < d(\mathcal{S}, \mathcal{T}_2)\).  
Define the quantile function \(Q_i(\delta)\) for the entropy loss of \(\boldsymbol{f}_{\boldsymbol{\theta}}\) on \(\mathcal{T}_i\) such that \(\mathbb{P}\left(H^\rho < \mathbb{E}[H^\rho] + Q_i(\delta)\right) = 1 - \delta\), and let \(Q(\delta) = \sup \{Q_1(\delta), Q_2(\delta)\}\) be the supremum quantile.  
Then, with probability at least \(1 - \delta\):  
\begin{align}
H^\rho(\boldsymbol{f}_{\boldsymbol{\theta}}(\boldsymbol{x}_{\scriptscriptstyle{\mathcal{T}_i}}))
\leq & \frac{M}{2} d\left(\mathcal{S}; \mathcal{T}_i\right) +Q(\delta/2)
+ \hat{H}^\rho_{\scriptscriptstyle{\mathcal{S}_n}}
+ 2\mu R_n(\mathcal{F}_v, \mathcal{S})
+ M \sqrt{\frac{\log (2 / \delta)}{2 n}} \nonumber \\
& + \log \frac{1}{\eta}\cdot \mathbb{E}_{\mathcal{S}} \|\mathbf{y}_{\scriptscriptstyle{\mathcal{S}}} - \boldsymbol{f}_{\widetilde{\boldsymbol{\theta}}}(\boldsymbol{x}_{\scriptscriptstyle{\mathcal{S}}})\|_1 
+ \frac{1}{\eta} \mathbb{E}_{\mathcal{S}} \|\mathbf{y}_{\scriptscriptstyle{\mathcal{S}}} - \boldsymbol{f}_{\widetilde{\boldsymbol{\theta}}}(\boldsymbol{x}_{\scriptscriptstyle{\mathcal{S}}})\|_1^2.
\end{align}
Here, the notation $\widetilde{\boldsymbol{\theta}}$ is defined as $\widetilde{\boldsymbol{\theta}} := \boldsymbol{\theta} + \arg\max_{\|\boldsymbol{\epsilon}\|\leq \rho} \max\left\{ H(\mathbf{y}, \boldsymbol{f}_{\boldsymbol{\theta}+\boldsymbol{\epsilon}}(\boldsymbol{x})), H(\boldsymbol{f}_{\boldsymbol{\theta}+\boldsymbol{\epsilon}}(\boldsymbol{x})) \right\}$. 
Furthermore, if this bound is \(\beta_i\)-tight for \(\mathcal{T}_1\) and \(\mathcal{T}_2\) with \(\beta_i < \gamma\), then there exists a threshold \(\xi\) such that: 
\begin{equation}
\mathbb{P}\left(H^\rho(\boldsymbol{f}_{{\boldsymbol{\theta}}}(\boldsymbol{x}_{\scriptscriptstyle{\mathcal{T}_1}})) < \xi\right) > \mathbb{P}\left(H^\rho(\boldsymbol{f}_{{\boldsymbol{\theta}}}(\boldsymbol{x}_{\scriptscriptstyle{\mathcal{T}_2}})) < \xi\right).
\end{equation}
\end{theorem}

This inequality indicates that a test distribution further from the training distribution tends to exhibit a higher sharpness score. By comparing sharpness scores across test distributions, we can identify which one is closer to the training distribution, thus yielding more reliable predictions. Notably, the tunable parameter $\rho$ controls the tightness of the upper bound, facilitating a precise differentiation between test distributions and improving the model performance.
It is important to note that in the theoretical analysis presented in this section, we do not distinguish between $\rho$ and $\rho'$ (which are utilized to calculate the sharpness of the training and test loss landscapes, respectively). However, in practical implementation, we may opt to use different values for $\rho$ and $\rho'$ for better performance.
Due to space limitations, detailed proofs and further discussions are provided in the appendices.

\section{Experiments}
\subsection{Experimental Setup}
\textbf{Datasets.} We conduct two types of experiments to evaluate the model's robustness to natural distribution shifts and its cross-dataset generalization capabilities, following previous research such as TPT~\citep{TPT}. To assess the model's robustness to natural distribution shifts, we apply prompt tuning on the ImageNet~\citep{ImageNet} dataset, and evaluate its performance on four ImageNet variants: ImageNet-A~\citep{ImageNet-A}, ImageNet-V2~\citep{ImageNet-V}, ImageNet-R~\citep{ImageNet-R} and ImageNet-Sketch~\citep{ImageNet-K}, which is also known as the domain generalization task. In addition, we perform cross-dataset evaluations for image classification across 10 datasets, each from a distinct domain with different classes: including Caltech101~\citep{caltech}, OxfordPets~\citep{pets}, StanfordCars~\citep{cars}, Flower102~\citep{flower}, Aircraft~\citep{aircraft}, SUN397~\citep{sun}, DTD~\citep{dtd},
Food101~\citep{food}, UCF101~\citep{ucf} and Eurosat~\citep{eurosat}.
In this experiment, ImageNet serves as the source dataset, while the remaining fine-grained datasets are used as target datasets for evaluation.

\textbf{Implementation details.} 
Our experiments are based on pretrained CLIP~\citep{CLIP} models, specifically CLIP-ResNet50 (using a ResNet50 image encoder) and CLIP-ViT-B/16 (using a Vision Transformer image encoder). 
Due to space limits, we focus on reporting the experimental results of CLIP-ViT-B/16, deferring those of CLIP-ResNet50 to the appendix.
In the prompt tuning stage, our experiments are built on the CoOp~\citep{coop} framework.
The prompts are trained in a 16-shot manner on the ImageNet dataset.
We set the number of prompts to 4 and utilize the SGD optimizer, with a learning rate of 0.002. 
For cross-dataset and domain generalization tasks, the prompts were trained for 5 and 50 epochs, with batch sizes of 4 and 32, respectively.
The key hyperparameters $\rho$ are determined through a grid search, with the values ranging from [0.05, 0.1, 0.3, 0.5, 0.7].
During testing, existing TPT-based methods usually leverage the input image along with its 63 augmented views. To ensure a fair comparison, we consistently employ the same data augmentation strategy as TPT in all experiments.
To avoid tuning hyperparameters on test data, we just set $\lambda = 1$ and $\rho'=0.5$ for all experiments. Please refer to the Appendix for more discussions about other hyperparameters.

\subsection{Main Results}

\setlength{\tabcolsep}{0.5mm}
\begin{table*}[t]
\centering
\caption{\textbf{Cross-dataset generalization from ImageNet to fine-grained classification datasets.} During the prompt tuning stage, the prompts are tuned on ImageNet with 16-shot training data per category, using a ViT-B/16 image encoder.}
\small
\scalebox{0.8}{\begin{tabular}{l|cccccccccc|c}
\toprule
Method & Caltech101 & Pets & Cars & Flowers102 & Aircraft & SUN397 & DTD & Eurosat & Food101 & UCF101 & Avg. \\
\midrule
CLIP-ViT-B/16~\citep{CLIP} & 93.35 & 88.25 & 65.48 & 67.44 & 23.67 & 62.59 & 44.27 & 42.01 & 83.65 & 65.13 & 63.58 \\
\midrule
CoOp~\citep{coop} & 93.70 & 89.14 & 64.51 & 68.71 & 18.47 & 64.15 & 41.92 & 46.39 & 85.30 & 66.55 & 63.88 \\
CoCoOp~\citep{cocoop} & {94.43} & 90.14 & 65.32 & 71.88 & 22.94 & 67.36 & 45.73 & 39.23 & 83.97 & {68.44} & 64.94 \\
\midrule
TPT~\citep{TPT} & 94.16 & 87.79 & 66.87 & 68.98 & 24.78 & 65.50 & \underline{47.75} & 42.44 & 84.67 & 68.04 & 65.10 \\
DiffTPT~\citep{DiffTPT}  & 92.49 & 88.22 & 67.01 & 70.10 & 25.60 & 65.74 & 47.00 & 43.13 & \textbf{87.23} & 62.67 & 64.92 \\
C-TPT~\citep{C-TPT} & 93.60 & 88.20 &  65.80 & 69.80  & 24.00 & 64.80 & 46.00 & 43.20 & 83.70 & 65.70 & 64.48 \\
ZERO~\citep{ZERO} & 93.66 & 87.75 & 68.04 & 67.68 & 25.21 & 65.03 & 46.12 & 34.33 & 86.53 & 67.77 & 64.21 \\
PromptAlign~\citep{align}  & 94.01 & \underline{90.76} & {68.50} & \textbf{72.39} & 24.80 & {67.54} & 47.24 & {47.86} & 86.65 & {69.47} & {66.92} \\
TDA~\citep{TDA} & 94.24 & 88.63 & 67.28 & 71.42 & 23.91 & 67.54 & 47.40 & \textbf{58.00} & 86.14 & \textbf{70.66} & \underline{67.53} \\
TPT+CoOp~\citep{coop} & 93.75 & 88.93 & 67.06 & 68.25 & {25.89} & 66.40 & 47.15 & \underline{48.78} & 83.82 & 66.53 & 65.66 \\
TPT+MaPLe~\citep{MaPLe} & 93.59 & 90.72 & 66.50 & \underline{72.37} & 24.70 & 67.54 & 45.87 & 47.80 & 86.64 & {69.19} & 66.50 \\
ZERO~\citep{ZERO}+CoOp & 93.85 & 88.36 & 64.90 & 67.23 & 19.14 & 64.73 & 43.62 & 33.53 & 82.67 & 66.61 & 62.46 \\
ZERO~\citep{ZERO}+MaPLe &  \underline{94.48} & 90.60 & \underline{68.58} & 71.62 & \underline{26.25} & \underline{68.20} & 45.86 & 42.17 & \underline{86.77} & \underline{69.87} & 66.44 \\
\midrule
\textbf{FGA (Ours)}  & \textbf{96.96} & \textbf{91.28} & \textbf{68.93} & 72.11 & \textbf{26.97} & \textbf{69.29} & \textbf{49.76} & {47.58} & {84.95} & {68.17} & \textbf{67.60} \\
\bottomrule
\end{tabular}}
\label{tab:cross-dataset}
\end{table*}

\textbf{Robustness to natural distribution shifts.} We first compare the proposed FGA with prevalent TTA techniques on ImageNet and its variant out-of-domain (OOD) datasets. The results, presented in Table \ref{tab:DG}, highlight the superior performance of FGA across several ImageNet-based OOD datasets. Notably, even the ablated version of our method, STSS+CoOp, exhibits strong performance, surpassing all previous approaches with an OOD average of 64.60\% and an overall average of 66.48\%. 
We attribute this robustness to the fact that standard SGD training already imbues models with an implicit bias toward flatter minima.
By explicitly enhancing this geometric property through SAPT, the full FGA algorithm achieves a substantial leap in generalization performance.
Specifically, 
when compared to TPT+CoOp, FGA shows an average accuracy improvement of 4.88\% (61.67\% $\rightarrow$ 66.55\%) on the OOD variants.
Furthermore, our FGA also consistently surpasses other powerful TTA methods (e.g., DiffTPT, C-TPT, ZERO, MTA, and SAR) when they are combined with CoOp. It is critical to note that while online TTA methods like SAR benefit from aggregating information across a test data stream, FGA operates in a more challenging single-sample adaptation setting. For a fair comparison, we have adapted SAR to the single-sample adaptation setting. These superior results of FGA strongly demonstrate its effectiveness in enhancing CLIP's out-of-domain generalization across diverse datasets.

\textbf{Cross-dataset generalization.}
We also observe superior performance of the FGA in evaluating cross-dataset generalization from ImageNet to various fine-grained classification benchmarks.
Based on the comprehensive results presented in Table 2, the proposed FGA method demonstrates superior overall performance, achieving the highest average accuracy of 67.60\% and attaining top-tier results on 6 out of 10 datasets, including a notably strong performance on Caltech101 (96.96\%). 
Furthermore, FGA (67.60\%) achieves an average accuracy improvement of 1.94\% over the powerful baseline TPT+CoOp (65.66\%). 
It also exhibits superior performance over the combinations of TTA methods (like TPT and ZERO) and different tuning methods (CoOp and MaPLe).
Please note that due to the significant difference in TTA settings, it is not imperative to expect single-sample TTA methods to surpass online TTA methods like TDA.
Overall, these superior results further validate its effectiveness in adapting to diverse datasets during testing. 
It is particularly valuable for VLMs like CLIP, as it enables models to recognize more fine-grained categories in image classification without the need for additional training.

\textbf{Runtime and memory efficiency.} To evaluate the practical deployment advantages of FGA, we quantify its computational overhead in terms of inference time per test image and peak GPU memory consumption. All experiments are conducted on a single NVIDIA Tesla V100 GPU under consistent settings. Specifically, FGA requires only 0.07s per image from ImageNet, which is 23.86× faster than DiffTPT (1.67s) and 8.86× faster than TPT (0.62s). In terms of memory usage, FGA consumes merely 4.14 GB, which is 4.67× lower than that of TPT (19.33GB). These results demonstrate that FGA achieves highly competitive test-time adaptation performance while maintaining notably low runtime and memory overhead, making it suitable for real-time and resource-constrained applications.

\vspace{-0.1cm}
\subsection{Ablation Study}
\textbf{Main components analysis.}  
Our ablation study on the domain generalization benchmark (using CLIP-ViT-B/16 architecture, shown in Table~\ref{tab:DG}) validates the necessity of each FGA component:  
(1) Sharpness-aware prompt tuning (SAPT) enhances generalization, boosting CoOp's average accuracy by 0.92\% (61.72\%$\rightarrow$62.64\%) on ImageNet and OOD datasets;  
(2) Test-time sharpness selection (STSS) drives major gains, with CoOp+STSS outperforming CoOp by 4.76\% (61.72\%$\rightarrow$66.48\%);  
(3) SAPT synergistically enhances STSS, where full FGA (CoOp+SAPT+STSS) achieves a 5.40\% gain over CoOp+SAPT (62.64\%$\rightarrow$68.04\%) (exceeding standalone STSS improvements of 4.76\%). This confirms that flatter minima from SAPT intrinsically improve test-time sample selection.  

\begin{figure*}[t] 
    \centering
    \begin{minipage}[b]{0.42\textwidth}
        \centering
        \includegraphics[width=\textwidth]{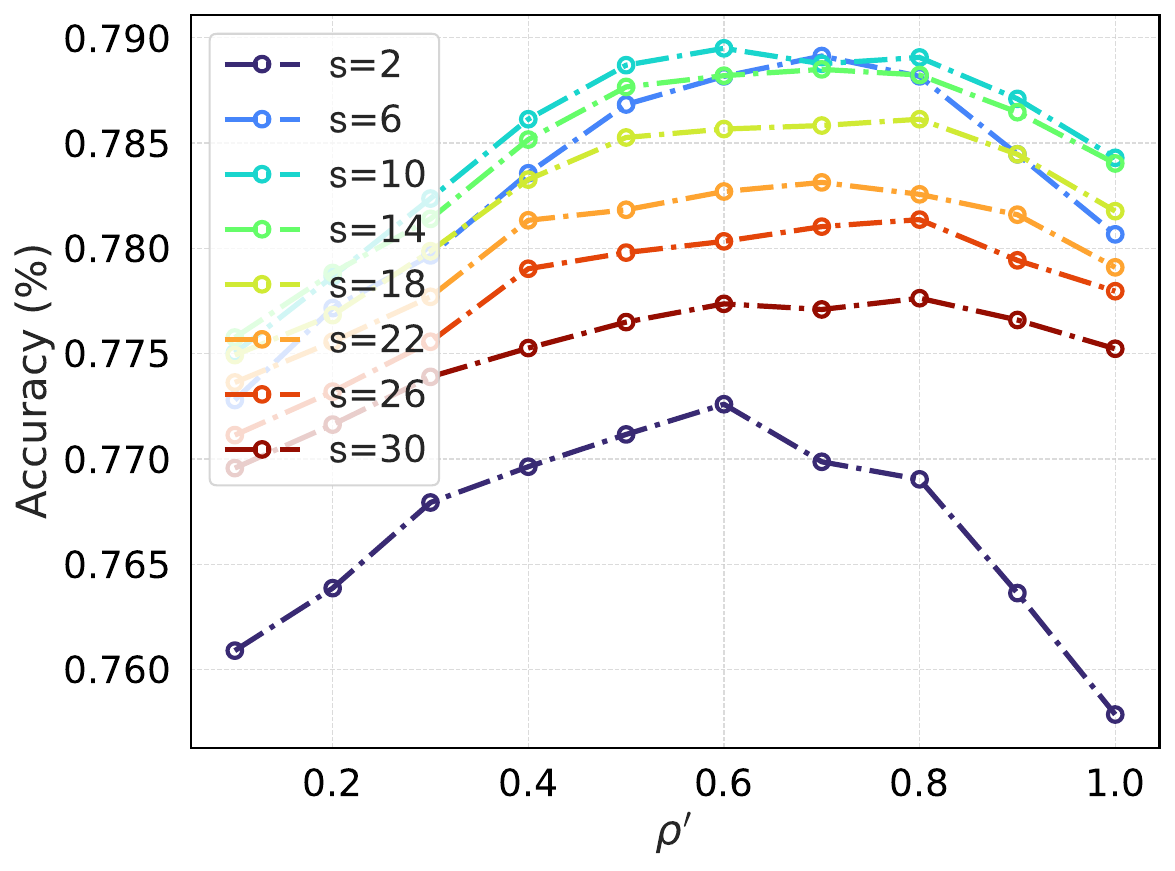}
        \caption{\textbf{The influence of the key hyperparameter $\rho'$ on the test accuracy.}} 
        \label{fig:ablation2}
    \end{minipage}
    \hfill
    \raisebox{0\height}{\begin{minipage}[b]{0.55\textwidth}  
        \centering
        \includegraphics[width=\textwidth]{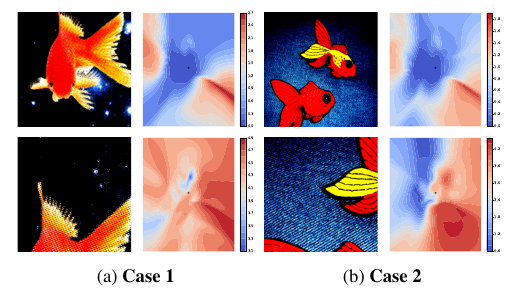}
        \caption{\textbf{2D Visualization of loss landscapes associated with different augmented test samples.}} 
        \label{fig:case}
    \end{minipage}}
\end{figure*}

\textbf{Ablative analysis on key parameter $\rho'$.} Theoretical analysis (Section 4) establishes $\rho$  and $\rho'$ as key generalization controllers, playing significant roles during the training and testing stages, respectively. Since $\rho$'s role has been well explored in previous research~\citep{SAM}, we focus on $\rho'$ for test-time adaptation: as mentioned earlier, it governs distribution distinguishability, with proper values enhancing prediction reliability through sensitive discrimination. Empirical validation on ImageNet-R (Figure \ref{fig:ablation2}) shows non-monotonic accuracy dependence on $\rho'$—initially rising then falling. It is because extreme values ($\rho'\to0$ or $\rho'\gg0$) may yield uninformative sharpness measures and degrade performance. Crucially, $\rho'=0$ degenerates to entropy maximization, and its comparison with non-zero cases also demonstrates sharpness's necessity. Notably, all experiments fix $\rho'=0.5$ without test-data tuning, and this analysis solely aims to demonstrate the control effect of $\rho'$ on generalization. Additionally, sample retention follows a similar trend: accuracy peaks then declines with increased retention. This reflects the probabilistic correlation: lower sharpness typically means greater proximity to the training distribution, meaning performance degrades when retaining excessively high-sharpness samples.

\subsection{Visualization of Loss Landscapes}
To intuitively validate FGA's effectiveness, we visualize the test data's loss surface using a 2D technique~\citep{LossVisual} in Figure \ref{fig:case}, revealing how sample selection enhances prediction reliability. The visualization demonstrates critical relationships: when parameters reside in flat minima (Figure \ref{fig:case}, top), augmented samples maintain semantic integrity and enable reliable predictions. Conversely, parameters outside flat minima (Figure \ref{fig:case}, bottom) yield distorted semantic representations that degrade generalization. This contrast clearly illustrates that FGA can help filter out unreliable test samples to mitigate their adverse effects, thereby strengthening the generalization ability of VLMs.

\section{Conclusion}

This paper demonstrates that flatness operates not just as a beneficial training characteristic but as a key geometric clue for test-time adaptation.
This understanding motivates the proposal of a novel framework, Flatness-Guided Adaptation (FGA), 
which utilizes the principle of loss landscape flatness as a unified guide to improve both training and test generalization against distribution shifts.
Different from previous TTA methods that often fine-tune prompts per sample, it directs adaptation by leveraging the geometric relationship between training minima and test-time loss landscapes. 
Specifically, it first identifies flat minima during prompt tuning and then ensures the alignment across training and test landscapes via a selective mechanism. Comprehensive experiments and theoretical analysis confirm FGA's effectiveness and superior performance. We anticipate this work will advance the understanding of loss landscapes and inspire future TTA technologies.

\textbf{Future work.} Our proposed FGA is grounded in a general analysis of loss landscape geometry, a foundational concept that is broadly applicable across diverse model architectures and learning paradigms. This foundation makes FGA a flexible component that could be integrated into modern visual-language models, advanced prompt tuning methods, or new types of test-time adaptation objectives to potentially enhance their performance.
In this work, we intentionally followed the experimental setup introduced in the TPT paper. This choice allows a controlled and fair comparison with prior TTA methods, helping us to clearly demonstrate the contribution of FGA itself. The extension of FGA to other experimental configurations would require extensive engineering efforts to conduct large-scale experiments across diverse settings, and thus remains our future work.

\section*{Acknowledgements}

This work was financially supported by the National Natural Science Foundation of China (NSFC) under Grant No.U20B2070.
The authors gratefully acknowledge the NSFC for its critical support of this research.
\clearpage
\bibliography{iclr2026_conference}
\bibliographystyle{iclr2026_conference}
\clearpage
\appendix

The appendices provide additional rigorous technical foundations and extended experimental validation. Section~\ref{secA} and \ref{secB} establishes the theoretical bedrock through complete proofs of Theorem 1 and Theorem 4 (stated in the main text), formally deriving the generalization error bounds via the divergence discrepancy and Rademacher complexity analysis while elucidating the roles of hyperparameters $\rho$ and $\rho'$ in loss landscape smoothing and test distribution discrimination. 
Section~\ref{secC} presents comprehensive experimental results on both domain generalization and cross-dataset benchmarks with CLIP-ResNet50 architecture (see Tables~\ref{tab:1}-\ref{tab:2}).
Then, Section~\ref{secD} delivers full reproducibility resources, specifying critical hyperparameters, code architecture, and hardware configurations to enable exact replication. 
Finally, the document ends with a note on LLM utilization (in Section~\ref{secE}) for language refinement.

\section{Proof of Theorem 1}\label{secA}

We begin by defining the divergence between two distributions, which is crucial for bounding the difference in their corresponding loss functions. 
\begin{definition}
For two probability distributions $\mathcal{S}$ and $\mathcal{T}$ defined on a common measurable space, the divergence $d\left(\mathcal{S}, \mathcal{T}\right)$ is defined as twice the supremum of the absolute difference between their probabilities over all measurable subsets $A$, i.e., 
\begin{equation}
d\left(\mathcal{S}, \mathcal{T}\right) := 2 \sup _{A} \left| \mathbb{P}_{\mathcal{S}}(A) - \mathbb{P}_{\mathcal{T}}(A) \right|,
\end{equation}
where $\mathbb{P}_{\mathcal{S}}(A)$ and $\mathbb{P}_{\mathcal{T}}(A)$ represent the probabilities of event $A$ under distributions $\mathcal{S}$ and $\mathcal{T}$, and $\sup_A$ denotes the supremum taken over all such subsets $A$. 
\end{definition}

With this definition in hand, we now move to a lemma that connects the distance between distributions to the difference in the losses, setting the stage for bounding the generalization error.

\begin{lemma}[\citep{SWAD}]\label{th:dist}
Consider an bounded loss function $\ell:\mathbb{R}^K \times \mathbb{R}\rightarrow [0,M]$. 
Given two distributions, $\mathcal{S}$ and $\mathcal{T}$, the difference between the loss with $\mathcal{S}$ and the loss with $\mathcal{T}$ is bounded by the
distance between $\mathcal{S}$ and $\mathcal{T}$:
\begin{equation}
\left|\mathbb{E}_\mathcal{T}\ell\left(\boldsymbol{f}_{\boldsymbol{\theta}} (\boldsymbol{x}_{\scriptscriptstyle{\mathcal{T}}}), y_{\scriptscriptstyle{\mathcal{T}}}\right)-
\mathbb{E}_\mathcal{S}\ell\left(\boldsymbol{f}_{\boldsymbol{\theta}} (\boldsymbol{x}_{\scriptscriptstyle{\mathcal{S}}}), y_{\scriptscriptstyle{\mathcal{S}}}\right)
\right| \leq \frac{M}{2} d\left(\mathcal{S} ; \mathcal{T}\right).
\end{equation}
\end{lemma}

\begin{proof}
Firstly, using the identity that for any non-negative random variable $X$, $\mathbb{E}[X] = \int_0^\infty \mathbb{P}(X > t)  dt$, we rewrite the difference as an integral over the tail probabilities of the loss:
\begin{align}
&\left|\mathbb{E}_\mathcal{T}\ell\left(\boldsymbol{f}_{\boldsymbol{\theta}} (\boldsymbol{x}_{\scriptscriptstyle{\mathcal{T}}}), y_{\scriptscriptstyle{\mathcal{T}}}\right)-
\mathbb{E}_\mathcal{S}\ell\left(\boldsymbol{f}_{\boldsymbol{\theta}} (\boldsymbol{x}_{\scriptscriptstyle{\mathcal{S}}}), y_{\scriptscriptstyle{\mathcal{S}}}\right)
\right| \nonumber \\
= &\left| \int_0^\infty \mathbb{P}_{\mathcal{T}}\left( \ell(\boldsymbol{f}_{\boldsymbol{\theta}}(\boldsymbol{x}_{\scriptscriptstyle{\mathcal{T}}}), y_{\scriptscriptstyle{\mathcal{T}}}) > t \right) dt - \int_0^\infty\mathbb{P}_{\mathcal{S}}\left( \ell(\boldsymbol{f}_{\boldsymbol{\theta}}(\boldsymbol{x}_{\scriptscriptstyle{\mathcal{S}}}), y_{\scriptscriptstyle{\mathcal{S}}}) > t \right) dt \right| \nonumber \\
\leq &\int_0^\infty \left| \mathbb{P}_{\mathcal{T}}\left( \ell(\boldsymbol{f}_{\boldsymbol{\theta}}(\boldsymbol{x}_{\scriptscriptstyle{\mathcal{T}}}), y_{\scriptscriptstyle{\mathcal{T}}}) > t \right) - \mathbb{P}_{\mathcal{S}}\left( \ell(\boldsymbol{f}_{\boldsymbol{\theta}}(\boldsymbol{x}_{\scriptscriptstyle{\mathcal{S}}}), y_{\scriptscriptstyle{\mathcal{S}}}) > t \right) \right|  dt.
\end{align}
Since the loss $\ell$ is bounded in $[0, M]$, the tail probability is zero for all $t > M$. Thus, the above integral simplifies to:
\begin{equation}
I = \int_0^M \left| \mathbb{P}_{\mathcal{T}}\left( \ell(\boldsymbol{f}_{\boldsymbol{\theta}}(\boldsymbol{x}_{\scriptscriptstyle{\mathcal{T}}}), y_{\scriptscriptstyle{\mathcal{T}}}) > t \right) - \mathbb{P}_{\mathcal{S}}\left( \ell(\boldsymbol{f}_{\boldsymbol{\theta}}(\boldsymbol{x}_{\scriptscriptstyle{\mathcal{S}}}), y_{\scriptscriptstyle{\mathcal{S}}}) > t \right) \right|  dt.
\end{equation}
This integral is bounded above by taking the supremum over all $t \in [0, M]$ and all hypotheses $\boldsymbol{f}_{\boldsymbol{\theta}}$:
\begin{equation}
I
\leq
M \cdot \sup_{t \in [0, M]} \sup_{\boldsymbol{f}_{\boldsymbol{\theta}}} \left| \mathbb{P}_{\mathcal{T}}\left( \ell(\boldsymbol{f}_{\boldsymbol{\theta}}(\boldsymbol{x}_{\scriptscriptstyle{\mathcal{T}}}), y_{\scriptscriptstyle{\mathcal{T}}}) > t \right) - \mathbb{P}_{\mathcal{S}}\left( \ell(\boldsymbol{f}_{\boldsymbol{\theta}}(\boldsymbol{x}_{\scriptscriptstyle{\mathcal{S}}}), y_{\scriptscriptstyle{\mathcal{S}}}) > t \right) \right|.
\end{equation}
Now, for each fixed $t$ and $\boldsymbol{f}_{\boldsymbol{\theta}}$, the set $\{ \ell(\boldsymbol{f}_{\boldsymbol{\theta}}(\boldsymbol{x}_{\scriptscriptstyle{\mathcal{S}}}), y_{\scriptscriptstyle{\mathcal{S}}}) > t \}$ is a measurable event; hence the double supremum over such sets is dominated by the supremum over all measurable events $A$. Consequently, we obtain:
\begin{equation}
\left|\mathbb{E}_\mathcal{T}\ell\left(\boldsymbol{f}_{\boldsymbol{\theta}} (\boldsymbol{x}_{\scriptscriptstyle{\mathcal{T}}}), y_{\scriptscriptstyle{\mathcal{T}}}\right)-
\mathbb{E}_\mathcal{S}\ell\left(\boldsymbol{f}_{\boldsymbol{\theta}} (\boldsymbol{x}_{\scriptscriptstyle{\mathcal{S}}}), y_{\scriptscriptstyle{\mathcal{S}}}\right)
\right|
\leq
M \cdot \sup_{A} \left| \mathbb{P}_{\mathcal{T}}(A) - \mathbb{P}_{\mathcal{S}}(A) \right|.
\end{equation}
By referring back to Definition 1, we recognize that this supremum equals half the divergence $d\left(\mathcal{S}, \mathcal{T}\right)$, which completes the proof.
\end{proof}

Next, we incorporate the Rademacher complexity bound for the generalization error. This bound will provide a crucial link between the expected loss and the sample complexity.

\begin{lemma}[Theorem 6.31~\citep{itbook}]\label{th:it}
Consider a real-valued function class $\mathcal{F}=\{f_{\boldsymbol{\theta}}(\cdot)\}$ and a bounded loss function $\ell: \mathbb{R} \times \mathbb{R} \to [0,M]$. Assume that the loss function
 $\ell(f, y)$ is $\mu$-Lipschitz with respect to $f$ :
\begin{equation}
\left|\ell(f, y)-\ell\left(f^{\prime}, y\right)\right| \leq \mu\left|f-f^{\prime}\right|.
\end{equation}
Let $\mathcal{S}_n=\{\boldsymbol{x}_i,y_i\}_{i=1}^{n}$ be $n$ i.i.d. samples from distribution $\mathcal{S}$. With probability at least $1-\delta$:
\begin{align}
&\mathbb{E}_{\mathcal{S}} \ell(f_{\boldsymbol{\theta}}(\boldsymbol{x}), y) 
\leq  \frac{1}{n} \sum_{i=1}^n \ell\left(f_{\boldsymbol{\theta}}\left(\boldsymbol{x}_i\right), y_i\right)  +2 \mu R_n(\mathcal{F}, \mathcal{S})+M \sqrt{\frac{\log (1 / \delta)}{2 n}}.
\end{align}
Here, $R_n(\mathcal{F}, \mathcal{S})$ represents the  expected Rademacher complexity:
\begin{align}
    R_n(\mathcal{F}, \mathcal{S})=\mathbb{E}_{(\boldsymbol{x}_i,y_i) \sim \mathcal{S}} \mathbb{E}_\sigma \sup _{f \in \mathcal{F}} \frac{1}{n} \sum_{i=1}^n \sigma_i f_{\boldsymbol{\theta}}(\boldsymbol{x}_i),
\end{align}
where $\sigma_1,\ldots,\sigma_n$ are independent uniform $\{\pm 1\}$-valued Bernoulli random variables.
\end{lemma}

The lemma above, which addresses real-valued function classes, can be readily extended to the setting of vector-valued function classes (Theorem~\ref{th:it-vec}). Before presenting its formal statement and proof, we first introduce a necessary supporting lemma (Lemma~\ref{cor:vec_rademacher}).

\begin{lemma}[Corollary 4~\citep{vec_rademacher}]\label{cor:vec_rademacher}
Let $\mathcal{X}$ be any set, $(\boldsymbol{x}_1, \dots, \boldsymbol{x}_n) \in \mathcal{X}^n$, let $\mathcal{F}_v$ be a class of functions $\boldsymbol{f} : \mathcal{X} \to \mathbb{R}^K$ and let $h_i : \mathbb{R}^K \to \mathbb{R}$ have Lipschitz norm $\mu$. Then
\begin{equation}
\mathbb{E} \sup_{\boldsymbol{f} \in F_v} \sum_{i=1}^n \sigma_i h_i \big( \boldsymbol{f}(\boldsymbol{x}_i) \big) \leq \sqrt{2} \mu \, \mathbb{E} \sup_{\boldsymbol{f} \in F_v} \sum_{i=1}^n \sum_{k} \sigma_{ik} f_k (x_i),
\end{equation}
where $\{\sigma_i\}$ and $\{\sigma_{ik}\}$ are independent Bernoulli sequences, and $f_k(\boldsymbol{x}_i)$ denotes the $k$-th coordinate of $\boldsymbol{f}(\boldsymbol{x}_i) \in \mathbb{R}^K$.
\end{lemma}

\begin{theorem}\label{th:it-vec}
Consider a vector-valued function class $\mathcal{F}_{v}=\{\boldsymbol{f}_{\boldsymbol{\theta}}(\cdot):\mathcal{X}\to \mathbb{R}^K \}$ and a bounded loss function $\ell: \mathbb{R}^K \times \mathcal{Y} \to [0,M]$. Assume that the loss function
 $\ell(\boldsymbol{f}, y)$ is $\mu$-Lipschitz with respect to $\boldsymbol{f}$.
Let $\mathcal{S}_n=\{\boldsymbol{x}_i,y_i\}_{i=1}^{n}$ be $n$ i.i.d. samples from distribution $\mathcal{S}$. With probability at least $1-\delta$:
\begin{align}
&\mathbb{E}_{\mathcal{S}} \ell(\boldsymbol{f}_{\boldsymbol{\theta}}(\boldsymbol{x}), y) 
\leq  \frac{1}{n} \sum_{i=1}^n \ell\left(f_{\boldsymbol{\theta}}\left(\boldsymbol{x}_i\right), y_i\right)  +2 \sqrt{2} \mu R_n(\mathcal{F}_v, \mathcal{S})+M \sqrt{\frac{\log (1 / \delta)}{2 n}}.
\end{align}
Here, $R_n(\mathcal{F}_v, \mathcal{S})$ represents the  expected Rademacher complexity:
\begin{align}
    R_n(\mathcal{F}_v, \mathcal{S})=\mathbb{E}_{(\boldsymbol{x}_i,y_i) \sim \mathcal{S}} \mathbb{E}_\sigma \sup _{\boldsymbol{f} \in \mathcal{F}_v} \frac{1}{n} \sum_{i=1}^n \boldsymbol{\sigma}_i^\top \boldsymbol{f}_{\boldsymbol{\theta}}(\boldsymbol{x}_i),
\end{align}
where $\boldsymbol{\sigma}_1,\ldots,\boldsymbol{\sigma}_n$ are independent uniform $\{\pm 1\}$-valued Bernoulli random vectors.
\end{theorem}
\begin{proof}
Considering the real-valued function class $\mathcal{G} = \{ g = \ell \circ \boldsymbol{f}_{\boldsymbol{\theta}} : \mathcal{X} \to \mathbb{R} \}$ and the identity map $h: \mathbb{R} \to \mathbb{R}$, which has Lipschitz norm $1$, we apply Lemma~\ref{th:it} to obtain that, with probability at least $1-\delta$,
\begin{equation}
\mathbb{E}_{\mathcal{S}}\bigl[ \ell(f_{\boldsymbol{\theta}}(\boldsymbol{x}), y) \bigr] 
\leq \frac{1}{n} \sum_{i=1}^n \ell\bigl(f_{\boldsymbol{\theta}}(\boldsymbol{x}_i), y_i\bigr) 
+ 2 R_n(\mathcal{G}, \mathcal{S}) 
+ M \sqrt{\frac{\log(1/\delta)}{2n}}, 
\end{equation}
where $R_n(\mathcal{G}, \mathcal{S})$ denotes the expected Rademacher complexity of $\mathcal{G}$ over $\mathcal{S}$:
\begin{align}
R_n(\mathcal{G}, \mathcal{S}) 
= \mathbb{E} \sup_{\ell\circ\boldsymbol{f} \in \mathcal{G}} \frac{1}{n} \sum_{i=1}^n \sigma_i \, \ell(\boldsymbol{f}_{\boldsymbol{\theta}}(\boldsymbol{x}_i), y_i) \leq \sqrt{2}\,\mu \;
\mathbb{E} \sup_{\boldsymbol{f}_{\boldsymbol{\theta}} \in \mathcal{F}} 
\frac{1}{n} \sum_{i=1}^n \boldsymbol{\sigma}_i^\top \boldsymbol{f}_{\boldsymbol{\theta}}(\boldsymbol{x}_i).
\end{align}
The last inequality follows from Lemma~\ref{cor:vec_rademacher}.
\end{proof}

Now, by combining Lemma~\ref{th:dist} and Lemma~\ref{th:it-vec}, we conclude that with probability at least $1-\delta$,
\begin{align}
\mathbb{E}_{\mathcal{T}}\bigl[ \ell^{\rho}(\boldsymbol{f}_{\boldsymbol{\theta}}(\boldsymbol{x}_{\scriptscriptstyle{\mathcal{T}}}), y_{\scriptscriptstyle{\mathcal{T}}}) \bigr] 
&\leq \frac{M}{2}\, d\left(\mathcal{S}; \mathcal{T}\right) 
+ \hat{\ell}^{\rho}_{\mathcal{S}_n}(\boldsymbol{f}_{\boldsymbol{\theta}}) 
+ 2\sqrt{2}\,\mu\, R_n(\mathcal{F}_v, \mathcal{S}) 
+ M \sqrt{\frac{\log(1/\delta)}{2n}},
\end{align}
where $\hat{\ell}^{\rho}_{\mathcal{S}_n}(\boldsymbol{f}_{\boldsymbol{\theta}}) := \frac{1}{n}\sum_{i=1}^n \ell^{\rho}\bigl(\boldsymbol{f}_{\boldsymbol{\theta}}(\boldsymbol{x}_i), y_i\bigr)$ denotes the empirical loss over the training set $\mathcal{S}_n$. This completes the proof of Theorem~1.

\textit{Remarks.} The generalization error bound above does not explicitly include \(\rho\), which could be a significant parameter for mitigating generalization error. However, it is important to note that \(\rho\) can enhance the smoothness of $\ell^{\rho}$, thereby potentially reducing the Lipschitz constant $\mu$, which is a contributing factor to the expected Rademacher complexity $R_n(\mathcal{F}_v, \mathcal{S})$.

\section{Proof of Theorem 4}\label{secB}

We begin by proving a simplified version of Theorem 4, stated as follows.

\begin{theorem}\label{th:main_simple}
Consider a vector-valued function class $\mathcal{F}_v = \{\boldsymbol{f}_{\boldsymbol{\theta}}(\cdot):\mathcal{X}\to\mathbb{R}^K\}$, where the parameters \(\boldsymbol{\theta}\) lie in a set ${\Theta}$ such that the loss function $\ell$ is bounded within \([0, M]\).  
Given a training distribution \(\mathcal{S}\) and two \(\gamma\)-separable test distributions \(\mathcal{T}_1\) and \(\mathcal{T}_2\), assume \(d(\mathcal{S}, \mathcal{T}_1) < d(\mathcal{S}, \mathcal{T}_2)\).  
Define the quantile function \(q_i(\delta)\) for the entropy loss of \(\boldsymbol{f}_{\boldsymbol{\theta}}\) on \(\mathcal{T}_i\) such that \(\mathbb{P}\left(H^\rho < \mathbb{E}[H^\rho] + q_i(\delta)\right) = 1 - \delta\), and let \(Q(\delta) = \sup \{Q_1(\delta), Q_2(\delta)\}\) be the supremum quantile.  
Then, with probability at least \(1 - \delta\):  
\begin{equation}   
\ell^\rho(\boldsymbol{f}_{\boldsymbol{\theta}}(\boldsymbol{x}_{\scriptscriptstyle{\mathcal{T}_i}}), y_{\scriptscriptstyle{\mathcal{T}_i}}) \leq \frac{M}{2} d(\mathcal{S}, \mathcal{T}_i) + \hat{\ell}^\rho_{\mathcal{S}}(\boldsymbol{f}_{\boldsymbol{\theta}}) + 2\sqrt{2}\,\mu\, R_n(\mathcal{F}_v, \mathcal{S}) + M \sqrt{\frac{\log(2/\delta)}{2n}} + Q(\delta/2).
\end{equation}
If this bound is \(\beta_i\)-tight for \(\mathcal{T}_1\) and \(\mathcal{T}_2\) with \(\beta_i < \gamma\), then there exists a threshold \(\xi\) such that:
\begin{equation}
\mathbb{P}\left(\ell^\rho(\boldsymbol{f}_{\boldsymbol{\theta}}(\boldsymbol{x}_{\scriptscriptstyle{\mathcal{T}_1}}), y_{\scriptscriptstyle{\mathcal{T}_1}}) < \xi\right) > \mathbb{P}\left(\ell^\rho(\boldsymbol{f}_{\boldsymbol{\theta}}(\boldsymbol{x}_{\scriptscriptstyle{\mathcal{T}_2}}), y_{\scriptscriptstyle{\mathcal{T}_2}}) < \xi\right).
\end{equation}
\end{theorem}

\begin{proof}
By Theorem 1, with probability at least $1-\delta/2$:
\begin{align}
&\mathbb{E}_{\mathcal{T}_i}[ \ell^{\rho}(\boldsymbol{f}_{\boldsymbol{\theta}}(\boldsymbol{x}_{\scriptscriptstyle{\mathcal{T}_i}}), y_{\scriptscriptstyle{\mathcal{T}_i}})]
\leq  \hat{\ell}^{\rho}_{\mathcal{S}_n}(\boldsymbol{f}_{\boldsymbol{\theta}}) + \frac{M}{2} d\left(\mathcal{S} ; \mathcal{T}_i\right) 
+2\sqrt{2}\,\mu\, R_n(\mathcal{F}_v, \mathcal{S})+M \sqrt{\frac{\log (2 / \delta)}{2 n}}.
\end{align}
Furthermore, by the definition of the quantile function, with probability at least $1-\delta/2$:
\begin{align}
&\ell^{\rho}(\boldsymbol{f}_{\boldsymbol{\theta}}(\boldsymbol{x}_{\scriptscriptstyle{\mathcal{T}_i}}), y_{\scriptscriptstyle{\mathcal{T}_i}}) 
\leq \mathbb{E}_{\mathcal{T}_i} [\ell^{\rho}(\boldsymbol{f}_{\boldsymbol{\theta}}(\boldsymbol{x}_{\scriptscriptstyle{\mathcal{T}_i}}), y_{\scriptscriptstyle{\mathcal{T}_i}})] + Q_i(\delta/2).
\end{align}
Combining these two results yields, with probability at least $1-\delta$:
\begin{equation}   
\ell^\rho(\boldsymbol{f}_{\boldsymbol{\theta}}(\boldsymbol{x}_{\scriptscriptstyle{\mathcal{T}_i}}), y_{\scriptscriptstyle{\mathcal{T}_i}}) \leq \frac{M}{2} d(\mathcal{S}, \mathcal{T}_i) + \hat{\ell}^\rho_{\mathcal{S}}(\boldsymbol{f}_{\boldsymbol{\theta}}) + 2\sqrt{2}\,\mu\, R_n(\mathcal{F}_v, \mathcal{S}) + M \sqrt{\frac{\log(2/\delta)}{2n}} + Q_i(\delta/2).
\end{equation}
Since $Q(\delta)=\sup\{Q_1(\delta),Q_2(\delta\}$ is the supremum quantile, it follows that
\begin{equation}   
\ell^\rho(\boldsymbol{f}_{\boldsymbol{\theta}}(\boldsymbol{x}_{\scriptscriptstyle{\mathcal{T}_i}}), y_{\scriptscriptstyle{\mathcal{T}_i}}) \leq \frac{M}{2} d(\mathcal{S}, \mathcal{T}_i) + \hat{\ell}^\rho_{\mathcal{S}}(\boldsymbol{f}_{\boldsymbol{\theta}}) + 2\sqrt{2}\,\mu\, R_n(\mathcal{F}_v, \mathcal{S}) + M \sqrt{\frac{\log(2/\delta)}{2n}} + Q(\delta/2).
\end{equation}

Then, for the subsequent analysis, we define two quantities, $\xi_1$ and $\xi_2$, as follows:
\begin{align}
    &\xi_{1,2}:= \frac{M}{2} d\left(\mathcal{S} ; \mathcal{T}_{1,2}\right)
+ Q(\delta/2) 
 +\hat{\ell}^\rho_{\mathcal{S}}(\boldsymbol{f}_{\boldsymbol{\theta}})
+2\sqrt{2}\,\mu\, R_n(\mathcal{F}_v, \mathcal{S})+M \sqrt{\frac{\log (2 / \delta)}{2 n}}. \label{eq:xi}
\end{align}
Next, we consider the relationship between the divergences \(d(\mathcal{S}; \mathcal{T}_1)\) and \(d(\mathcal{S}; \mathcal{T}_2)\). If \(d(\mathcal{S}; \mathcal{T}_1) < d(\mathcal{S}; \mathcal{T}_2)\), then we immediately conclude that $\xi_1<\xi_2$.
From Theorem 1, we can further derive the following two inequalities:
\begin{align}
\mathbb{P}\left(\ell^{\rho}(\boldsymbol{f}_{\boldsymbol{\theta}}(\boldsymbol{x}_{\scriptscriptstyle{\mathcal{T}_1}}),y_{\scriptscriptstyle{\mathcal{T}_1}}) > \xi_1\right) < \delta,
\end{align}
and
\begin{align}
\mathbb{P}\left(\ell^{\rho}(\boldsymbol{f}_{\boldsymbol{\theta}}(\boldsymbol{x}_{\scriptscriptstyle{\mathcal{T}_2}}),y_{\scriptscriptstyle{\mathcal{T}_2}}) > \xi_2\right) < \delta.
\end{align}
To proceed, let us introduce $\xi^\star$, the oracle upper bound of $\ell^{\rho}(\boldsymbol{f}_{\boldsymbol{\theta}}(\boldsymbol{x}_{\scriptscriptstyle{\mathcal{T}_2}}),y_{\scriptscriptstyle{\mathcal{T}_2}})$, which satisfies:
\begin{align}
    \mathbb{P}\left(\ell^{\rho}(\boldsymbol{f}_{\boldsymbol{\theta}}(\boldsymbol{x}_{\scriptscriptstyle{\mathcal{T}_2}}),y_{\scriptscriptstyle{\mathcal{T}_2}}) > \xi^\star\right) = \delta.
\end{align}
Next, we define the difference $\beta$ as the absolute difference between $\xi_2$ and the oracle bound $\xi^\star$, i.e., $\beta:=|\xi_2-\xi^\star|$.
Given the separability of $\mathcal{T}_1$ and $\mathcal{T}_2$, we can infer that $|\xi_2 - \xi_1| > \beta$. This implies that: $\xi_1<\xi_2-\beta\leq \xi^\star$.
Consequently, we can conclude that:
\begin{align}
    \mathbb{P}\left(\ell^{\rho}(\boldsymbol{f}_{\boldsymbol{\theta}}(\boldsymbol{x}_{\scriptscriptstyle{\mathcal{T}_2}}),y_{\scriptscriptstyle{\mathcal{T}_2}}) >\xi_1\right)
     > \mathbb{P}\left(\ell^{\rho}(\boldsymbol{f}_{\boldsymbol{\theta}}(\boldsymbol{x}_{\scriptscriptstyle{\mathcal{T}_2}}),y_{\scriptscriptstyle{\mathcal{T}_2}})>\xi^\star\right).
\end{align}
Now, let us set $\xi = \xi_1$, which leads us to the following inequalities:
\begin{align}
    \mathbb{P}\left(\ell^{\rho}(\boldsymbol{f}_{\boldsymbol{\theta}}(\boldsymbol{x}_{\scriptscriptstyle{\mathcal{T}_1}}),y_{\scriptscriptstyle{\mathcal{T}_1}}) > \xi\right) < \delta < \mathbb{P}\left(\ell^{\rho}(\boldsymbol{f}_{\boldsymbol{\theta}}(\boldsymbol{x}_{\scriptscriptstyle{\mathcal{T}_2}}),y_{\scriptscriptstyle{\mathcal{T}_2}}) > \xi_1\right),
\end{align}
which completes the proof of Theorem~\ref{th:main_simple}.
\end{proof}

Please note that in the absence of true labels during test-time adaptation, we must employ entropy as a surrogate loss for cross entropy, which is equivalent to using cross entropy with conjugate pseudo-labels and is considered the best alternative to cross entropy\citep{conjy}. However, using a surrogate loss introduces an additional error term, which relates to the following theorem.

\begin{theorem}[Upper bound on the difference between entropy and cross entropy] \label{th:ent}
Let \( \boldsymbol{p} = (p_1, \dots, p_K) \) and \( \boldsymbol{q} = (q_1, \dots, q_K) \) be two probability distributions over a finite set \( \{1, \dots, K\} \). Suppose there exists a constant \( \eta > 0 \) such that \( p_i,q_i \geq \eta \) for all \( i \). Then the entropy \( H(\boldsymbol{q}) \) and the cross entropy \( H(\boldsymbol{p}, \boldsymbol{q}) \) satisfy the following inequality:

\begin{equation}
|H(\boldsymbol{p},\boldsymbol{q}) - H(\boldsymbol{q})| \leq \log\frac{1}{\eta}  \cdot \|\boldsymbol{p} - \boldsymbol{q}\|_1 + \frac{1}{\eta} \|\boldsymbol{p} - \boldsymbol{q}\|_1^2,
\end{equation}

where \( \|\boldsymbol{p} - \boldsymbol{q}\|_1 = \sum_{i=1}^K |p_i - q_i| \) is the \( L^1 \) distance between the distributions.
\end{theorem}
\begin{proof}
We begin by recalling the relevant definitions. The entropy of \( q \) is \( H(\boldsymbol{q}) = -\sum_{i=1}^K q_i \log q_i \), and the cross entropy between \( p \) and \( q \) is \( H(\boldsymbol{p},\boldsymbol{q}) = -\sum_{i=1}^K p_i \log q_i \). The Kullback–Leibler divergence is defined as \( D_{\text{KL}}(\boldsymbol{p} \| \boldsymbol{q}) = \sum_{i=1}^K p_i \log \frac{p_i}{q_i} \). A well-known identity relates these quantities:  
\begin{equation}
H(\boldsymbol{p},\boldsymbol{q}) = H(\boldsymbol{p}) + D_{\text{KL}}(\boldsymbol{p} \| \boldsymbol{q}).
\end{equation}  
We can express the difference of interest as:  
\begin{equation}
H(\boldsymbol{p},\boldsymbol{q}) - H(\boldsymbol{q}) = \left[ H(\boldsymbol{p},\boldsymbol{q}) - H(\boldsymbol{p}) \right] + \left[ H(\boldsymbol{p}) - H(\boldsymbol{q}) \right] = D_{\text{KL}}(\boldsymbol{p} \| \boldsymbol{q}) + \left[ H(\boldsymbol{p}) - H(\boldsymbol{q}) \right].
\end{equation}  
Taking absolute values and applying the triangle inequality yields:  
\begin{equation}
|H(\boldsymbol{p},\boldsymbol{q}) - H(\boldsymbol{q})| \leq D_{\text{KL}}(\boldsymbol{p} \| \boldsymbol{q}) + |H(\boldsymbol{p}) - H(\boldsymbol{q})|. \label{eq:1}
\end{equation}  
The rest of the proof consists of bounding the two terms on the right-hand side.

First, we bound the KL divergence.
We establish the upper bound for the KL divergence using a second-order Taylor expansion along the linear path connecting the two distributions. Define the parametric family of distributions \( \boldsymbol{r}(t) = (1-t)\boldsymbol{q} + t \boldsymbol{p} \) for \( t \in [0, 1] \), which interpolates between \( \boldsymbol{r}(0) = \boldsymbol{q} \) and \( \boldsymbol{r}(1) = \boldsymbol{p} \). Consider the function \( f(t) = D_{\mathrm{KL}}(\boldsymbol{r}(t) \| \boldsymbol{q}) \), which measures the KL divergence from the interpolated distribution to the target distribution \( \boldsymbol{q} \). Our goal is to compute \( f(1) = D_{\mathrm{KL}}(\boldsymbol{p} \| \boldsymbol{q}) \) using Taylor's theorem.
We begin by computing the derivatives of \( f(t) \):
\begin{equation}
f'(t) = \sum_i (p_i - q_i) \log \frac{r_i(t)}{q_i},\quad  f''(t) = \sum_i \frac{(p_i - q_i)^2}{r_i(t)}.
\end{equation}
We now apply Taylor's theorem with Lagrange remainder at \( t = 0 \). Since \( f(0) = D_{\mathrm{KL}}(\boldsymbol{q} \| \boldsymbol{q} ) = 0 \) and \( f'(0) = \sum_i (p_i - q_i) \log \frac{q_i}{q_i} = 0 \), the expansion yields \( f(t) = \frac{1}{2} f''(s) t^2 \) for some \( s \in [0, t] \). Setting \( t = 1 \) gives the key identity \( D_{\mathrm{KL}}(\boldsymbol{p} \| \boldsymbol{q}) = \frac{1}{2} \sum_i \frac{(p_i - q_i)^2}{r_i(s)} \), where \( r_i(s) = (1-s)q_i + s p_i \).
The critical step is to bound the denominator \( r_i(s) \) away from zero. Since both \( p_i \geq \eta \) and \( q_i \geq \eta \) by assumption, and since \( r_i(s) \) is a convex combination of \( p_i \) and \( q_i \), we have \( r_i(s) \geq \eta \) for all \( i \) and all \( s \in [0, 1] \). This allows us to bound each term in the sum by \( \frac{(p_i - q_i)^2}{r_i(s)} \leq \frac{(p_i - q_i)^2}{\eta} \). Summing over all components gives \( \sum_i \frac{(p_i - q_i)^2}{r_i(s)} \leq \frac{1}{\eta} \sum_i (p_i - q_i)^2 \).
Then, we relate the sum of squares to the \( L^1 \) distance. 
For each \( i \), we have \( (p_i - q_i)^2 \leq |p_i - q_i| \cdot \| \boldsymbol{p} - \boldsymbol{q} \|_1 \) because \( |p_i - q_i| \leq \| \boldsymbol{p} - \boldsymbol{q} \|_1 \). Summing this inequality over \( i \) gives \( \sum_i (p_i - q_i)^2 \leq \| \boldsymbol{p} - \boldsymbol{q} \|_1^2 \).
Combining this with the previous bound, we obtain:
\begin{equation}
 D_{\mathrm{KL}}(\boldsymbol{p} \| \boldsymbol{q}) \leq \frac{1}{2\eta} \| \boldsymbol{p} - \boldsymbol{q} \|_1^2. \label{eq:2}
\end{equation}

Next, we bound the difference \( |H(\boldsymbol{p}) - H(\boldsymbol{q})| \). 
Consider the entropy function \( H(\boldsymbol{r}) = -\sum_{i=1}^K r_i \log r_i \), defined on the probability simplex. Its gradient has components \( \frac{\partial H}{\partial r_i} = -\log r_i - 1 \), and its Hessian is a diagonal matrix with entries \( -\frac{1}{r_i} \) on the diagonal. We perform a second-order Taylor expansion of \( H(\boldsymbol{q}) \) around \( \boldsymbol{p} \):  
\begin{equation}
H(\boldsymbol{q}) = H(\boldsymbol{p}) + \langle \nabla H(\boldsymbol{p}), \boldsymbol{q} - \boldsymbol{p}  \rangle + \frac{1}{2} (\boldsymbol{q} - \boldsymbol{p} )^\top \nabla^2 H(\boldsymbol{\xi})(\boldsymbol{q} - \boldsymbol{p} ),
\end{equation}  
where \( \boldsymbol{\xi} \) lies on the segment between \( \boldsymbol{p} \) and \( \boldsymbol{q} \). Since \( p_i,q_i \geq \eta \) and the simplex is convex, we also have \( \xi_i \geq \eta \). We now bound each term.
The linear term satisfies:  
\begin{equation}
| \langle \nabla H(\boldsymbol{p}), \boldsymbol{q} - \boldsymbol{p}  \rangle | = | \langle \nabla \Tilde{H}(\boldsymbol{p}), \boldsymbol{q} - \boldsymbol{p}  \rangle |\leq \| \nabla \Tilde{H}(\boldsymbol{p}) \|_\infty \cdot \|\boldsymbol{q} - \boldsymbol{p} \|_1.
\end{equation} 
Here, since $\sum_i (q_i-p_i) = 0$, we use the notation $\nabla \Tilde{H}(\boldsymbol{p})$ to denote $\nabla H(\boldsymbol{p}) + 1$.
Because $\| \nabla \Tilde{H}(\boldsymbol{p}) \|_\infty = \max_{1 \leq j \leq K} | \log p_j  | =\max_{1 \leq j \leq K} | \log \frac{1}{p_j}| \leq \log \frac{1}{\eta}$, we have
\begin{equation}
| \langle \nabla H(\boldsymbol{p}), \boldsymbol{q} - \boldsymbol{p}  \rangle | \leq  \log \frac{1}{\eta} \cdot \|\boldsymbol{p} - \boldsymbol{q}\|_1. \label{eq:3}
\end{equation}  
For the quadratic term, the Hessian satisfies \( |(\boldsymbol{q} - \boldsymbol{p} )^\top \nabla^2 H(\boldsymbol{\xi})(\boldsymbol{q} - \boldsymbol{p} )| \leq \frac{1}{\eta} \|\boldsymbol{q} - \boldsymbol{p} \|_2^2 \). Since \( \|\boldsymbol{q} - \boldsymbol{p} \|_2^2 \leq \|\boldsymbol{q} - \boldsymbol{p} \|_1^2 \), we obtain:  
\begin{equation}
\left| \frac{1}{2} (\boldsymbol{q} - \boldsymbol{p} )^\top \nabla^2 H(\boldsymbol{\xi})(\boldsymbol{q} - \boldsymbol{p} ) \right| \leq \frac{1}{2\eta} \|\boldsymbol{p} - \boldsymbol{q}\|_1^2.\label{eq:4}
\end{equation}  
Combining (\ref{eq:3}) and (\ref{eq:4}), we conclude:  
\begin{equation}
|H(\boldsymbol{p}) - H(\boldsymbol{q})| \leq \log \frac{1}{\eta} \cdot \|\boldsymbol{p} - \boldsymbol{q}\|_1 + \frac{1}{2\eta} \|\boldsymbol{p} - \boldsymbol{q}\|_1^2. \label{eq:5}
\end{equation}

Substituting the bounds (\ref{eq:2}) and (\ref{eq:5}) into inequality (\ref{eq:1}) gives:  
\begin{equation}
|H(\boldsymbol{p},\boldsymbol{q}) - H(\boldsymbol{q})| \leq \log \frac{1}{\eta} \cdot \|\boldsymbol{p} - \boldsymbol{q}\|_1 + \frac{1}{\eta} \|\boldsymbol{p} - \boldsymbol{q}\|_1^2,
\end{equation}  
which completes the proof.
\end{proof} 

A direct application of Theorem~\ref{th:main_simple} and Theorem~\ref{th:ent} yields the following corollary, which is formally stated as Theorem 4.

\begin{corollary}
Consider a vector-valued function class $\mathcal{F}_v = \{\boldsymbol{f}_{\boldsymbol{\theta}}(\cdot):\mathcal{X}\to\mathbb{R}^K\}$, where the parameters \(\boldsymbol{\theta}\) lie in a set ${\Theta}$ such that the loss function is bounded within \([0, M]\).  
Let \(\boldsymbol{f} = (f_1, \dots, f_K)\) and \(\mathbf{y} = (\mathrm{y}_1, \dots, \mathrm{y}_K)\) be probability distributions over a finite set \(\{1, \dots, K\}\), with \(f_i, \mathrm{y}_i \geq \eta > 0\) for all \(i\). Denote by \(H(\boldsymbol{f})\) the entropy and by \(H(\mathbf{y}, \boldsymbol{f})\) the cross entropy.  Assume the worst-case cross entropy $H^{\rho}(\mathbf{y},\boldsymbol{f})$ is $\mu$-Lipschitz continuous with respect to $\boldsymbol{f}$ over ${\Theta}$.
Given a training distribution \(\mathcal{S}\) and two \(\gamma\)-separable test distributions \(\mathcal{T}_1\) and \(\mathcal{T}_2\), assume \(d(\mathcal{S}, \mathcal{T}_1) < d(\mathcal{S}, \mathcal{T}_2)\).  
Define the quantile function \(Q_i(\delta)\) for the entropy loss of \(\boldsymbol{f}_{\boldsymbol{\theta}}\) on \(\mathcal{T}_i\) such that \(\mathbb{P}\left(H^\rho < \mathbb{E}[H^\rho] + Q_i(\delta)\right) = 1 - \delta\), and let \(Q(\delta) = \sup \{Q_1(\delta), Q_2(\delta)\}\) be the supremum quantile.  
Then, with probability at least \(1 - \delta\):  
\begin{align}
H^\rho(\boldsymbol{f}_{\boldsymbol{\theta}}(\boldsymbol{x}_{\scriptscriptstyle{\mathcal{T}_i}}))
\leq & \frac{M}{2} d\left(\mathcal{S}; \mathcal{T}_i\right) +Q(\delta/2)
+ \hat{H}^\rho_{\scriptscriptstyle{\mathcal{S}_n}}
+ 2\mu R_n(\mathcal{F}_v, \mathcal{S})
+ M \sqrt{\frac{\log (2 / \delta)}{2 n}} \nonumber \\
& + \log \frac{1}{\eta}\cdot \mathbb{E}_{\mathcal{S}} \|\mathbf{y}_{\scriptscriptstyle{\mathcal{S}}} - \boldsymbol{f}_{\widetilde{\boldsymbol{\theta}}}(\boldsymbol{x}_{\scriptscriptstyle{\mathcal{S}}})\|_1 
+ \frac{1}{\eta} \mathbb{E}_{\mathcal{S}} \|\mathbf{y}_{\scriptscriptstyle{\mathcal{S}}} - \boldsymbol{f}_{\widetilde{\boldsymbol{\theta}}}(\boldsymbol{x}_{\scriptscriptstyle{\mathcal{S}}})\|_1^2.
\end{align}
Here, the notation $\widetilde{\boldsymbol{\theta}}$ is defined as $\widetilde{\boldsymbol{\theta}} := \boldsymbol{\theta} + \arg\max_{\|\boldsymbol{\epsilon}\|\leq \rho} \max\left\{ H(\mathbf{y}, \boldsymbol{f}_{\boldsymbol{\theta}+\boldsymbol{\epsilon}}(\boldsymbol{x})), H(\boldsymbol{f}_{\boldsymbol{\theta}+\boldsymbol{\epsilon}}(\boldsymbol{x})) \right\}$. 
Furthermore, if this bound is \(\beta_i\)-tight for \(\mathcal{T}_1\) and \(\mathcal{T}_2\) with \(\beta_i < \gamma\), then there exists a threshold \(\xi\) such that: 
\begin{equation}
\mathbb{P}\left(H^\rho(\boldsymbol{f}_{{\boldsymbol{\theta}}}(\boldsymbol{x}_{\scriptscriptstyle{\mathcal{T}_1}})) < \xi\right) > \mathbb{P}\left(H^\rho(\boldsymbol{f}_{{\boldsymbol{\theta}}}(\boldsymbol{x}_{\scriptscriptstyle{\mathcal{T}_2}})) < \xi\right).
\end{equation}
\end{corollary}
\begin{proof}
We begin by extending the conclusion of Theorem~\ref{th:ent} to the sharpness-aware entropy $H^\rho$. Specifically, we establish a bound for the difference between the sharpness-aware cross-entropy and entropy. By the definition of sharpness-ware loss, we have:
\begin{equation}
\left| H^{\rho}(\mathbf{y}, \boldsymbol{f}_{\boldsymbol{\theta}}(\boldsymbol{x})) - H^\rho(\boldsymbol{f}_{\boldsymbol{\theta}}(\boldsymbol{x})) \right| = \left| H(\mathbf{y},\boldsymbol{f}_{\boldsymbol{\theta}'}(\boldsymbol{x})) - H(\boldsymbol{f}_{\boldsymbol{\theta}''}(\boldsymbol{x})) \right|,
\end{equation}
where the perturbed parameters $\boldsymbol{\theta}'$ and $\boldsymbol{\theta}''$ are defined as:
\begin{align}
\boldsymbol{\theta}' &= \boldsymbol{\theta} + \arg\max_{\|\boldsymbol{\epsilon}\|\leq \rho} H(\mathbf{y}, \boldsymbol{f}_{\boldsymbol{\theta}+\boldsymbol{\epsilon}}(\boldsymbol{x})), \\
\boldsymbol{\theta}'' &= \boldsymbol{\theta} + \arg\max_{\|\boldsymbol{\epsilon}\|\leq \rho} H(\boldsymbol{f}_{\boldsymbol{\theta}+\boldsymbol{\epsilon}}(\boldsymbol{x})).
\end{align}
To establish a unified framework, we introduce a combined parameter $\widetilde{\boldsymbol{\theta}}$ defined by:
\begin{equation}
\widetilde{\boldsymbol{\theta}} = \boldsymbol{\theta} + \arg\max_{\|\boldsymbol{\epsilon}\|\leq \rho} \max\left\{ H(\mathbf{y}, \boldsymbol{f}_{\boldsymbol{\theta}+\boldsymbol{\epsilon}}(\boldsymbol{x})), H(\boldsymbol{f}_{\boldsymbol{\theta}+\boldsymbol{\epsilon}}(\boldsymbol{x})) \right\}.
\end{equation}
This unified parameter allows us to bound the difference between the sharpness-aware entropies through a single reference point.
Applying Theorem~\ref{th:ent}, we obtain the following inequality:
\begin{align}
\left| H^{\rho}(\mathbf{y}, \boldsymbol{f}_{\boldsymbol{\theta}}(\boldsymbol{x})) - H^\rho(\boldsymbol{f}_{\boldsymbol{\theta}}(\boldsymbol{x})) \right| 
&= \left| H(\mathbf{y},\boldsymbol{f}_{\boldsymbol{\theta}'}(\boldsymbol{x})) - H(\boldsymbol{f}_{\boldsymbol{\theta}''}(\boldsymbol{x})) \right| \nonumber \\
&\leq \left| H(\mathbf{y},f_{\widetilde{\boldsymbol{\theta}}}(\boldsymbol{x})) - H(f_{\widetilde{\boldsymbol{\theta}}}(\boldsymbol{x})) \right| \nonumber \\
&\leq \log \frac{1}{\eta} \cdot \|\mathbf{y} - f_{\widetilde{\boldsymbol{\theta}}}(\boldsymbol{x})\|_1 + \frac{1}{\eta} \|\mathbf{y} - f_{\widetilde{\boldsymbol{\theta}}}(\boldsymbol{x})\|_1^2.
\end{align}
Taking expectations over the training distribution $\mathcal{S}$, we derive:
\begin{align}
\mathbb{E}_{\mathcal{S}} \left[ H^\rho(\boldsymbol{f}_{\boldsymbol{\theta}}(\boldsymbol{x})) \right] 
\leq \mathbb{E}_{\mathcal{S}} \left[ H^\rho(\mathbf{y}, \boldsymbol{f}_{\boldsymbol{\theta}}(\boldsymbol{x})) \right] 
+ \log \frac{1}{\eta} \cdot \mathbb{E}_{\mathcal{S}} \|\mathbf{y} - f_{\widetilde{\boldsymbol{\theta}}}(\boldsymbol{x})\|_1 
+ \frac{1}{\eta} \mathbb{E}_{\mathcal{S}} \|\mathbf{y} - f_{\widetilde{\boldsymbol{\theta}}}(\boldsymbol{x})\|_1^2.
\end{align}
Finally, by applying Lemma~\ref{th:dist} and Lemma~\ref{th:ent}, we obtain the complete generalization bound for the sharpness-aware entropy on the target distribution $\mathcal{T}$:
\begin{align}
\mathbb{E}_{\mathcal{T}} \left[ H^\rho(\boldsymbol{f}_{\boldsymbol{\theta}}(\boldsymbol{x}_{\scriptscriptstyle{\mathcal{T}}})) \right]
\leq & \frac{M}{2} d\left(\mathcal{S}; \mathcal{T}\right)
+ \hat{H}^\rho_{\scriptscriptstyle{\mathcal{S}_n}}
+ 2\sqrt{2}\,\mu\, R_n(\mathcal{F}_v, \mathcal{S})
+ M \sqrt{\frac{\log (1 / \delta)}{2 n}} \nonumber \\
& + \log \frac{1}{\eta}\cdot \mathbb{E}_{\mathcal{S}} \|\mathbf{y}_{\scriptscriptstyle{\mathcal{S}}} - f_{\widetilde{\boldsymbol{\theta}}}(\boldsymbol{x}_{\scriptscriptstyle{\mathcal{S}}})\|_1 
+ \frac{1}{\eta} \mathbb{E}_{\mathcal{S}} \|\mathbf{y}_{\scriptscriptstyle{\mathcal{S}}} - \boldsymbol{f}_{\widetilde{\boldsymbol{\theta}}}(\boldsymbol{x}_{\scriptscriptstyle{\mathcal{S}}})\|_1^2.
\end{align}

This establishes an upper bound for the sharpness-aware entropy, completing the extension of Theorem~\ref{th:main_simple} to $H^\rho$. The remainder of the proof follows without modification, as the derived bound maintains the same structural properties as the original theorem while incorporating the sharpness-aware formulation.
\end{proof}

\section{Results of CLIP-ResNet50}\label{secC}

The experimental results with the CLIP-ResNet50 also demonstrate FGA's significant advancements in vision-language generalization across both natural distribution shifts (Table~\ref{tab:DG_RN50}) and fine-grained cross-dataset adaptation (Table~\ref{tab:cross-dataset-RN50}). 

In Table 1's natural shift evaluation, FGA achieves state-of-the-art performance with 52.55\% average accuracy, outperforming the prior best method (DiffTPT+CoOp) by +1.68\%. Notably, it shows remarkable gains on challenging out-of-distribution (OOD) benchmarks, particularly in ImageNet-A (37.87\%), where it improves upon DiffTPT+CoOp by +4.91\%, highlighting exceptional robustness to natural distribution shifts.

Table 2 further reveals FGA's superiority in fine-grained adaptation, achieving a record 62.56\% average accuracy that exceeds the previous best (DiffTPT) by +0.67\%. 
These results further validate the effectiveness of the FGA in adapting to diverse datasets during testing. 
We observe substantial improvements over baselines: +2.36\% on Caltech101 (vs. CoCoOp) and +2.14\% on Aircraft (vs. DiffTPT), highlighting the effectiveness of FGA in challenging fine-grained domains.

The consistent top-tier performance across both benchmarks, particularly on fine-grained datasets like Caltech101 and distributionally-shifted datasets like ImageNet-A, confirms FGA's ability to simultaneously enhance generalization robustness and fine-grained recognition precision.

\setlength{\tabcolsep}{2mm}
\begin{table*}[t]
\centering
\small
\caption{\textbf{Results with CLIP-ResNet50 on the domain generalization benchmark.} We report
top-1 accuracy (\%) for each method across five target datasets, using the CLIP-ResNet50 backbone. We highlight the best results in \textbf{bold} and \underline{underline} the second best results.
}\label{tab:1}
\scalebox{0.95}{\begin{tabular}{l|ccccc|c|c}
\toprule
Algorithm & IN & IN-A & IN-V2 & IN-R & IN-Sketch & Avg. & OOD Avg. \\

\midrule
CLIP-ResNet50~\citep{CLIP} &  59.81 & 23.24 & 52.91 & \underline{60.72} & 35.48 & 46.43 & 43.09 \\
\midrule
CoOp~\citep{coop} & 63.33 & 23.06 & 55.40 & 56.60 & 34.67 & 46.61 & 42.43 \\
CoCoOp~\citep{cocoop} & 62.81 & 23.32 & 55.72 & 57.74 & 34.48 & 46.81 & 42.82 \\
Tip-Adapter~\citep{tip} & 62.03 & 23.13 & 53.97 & 60.35 & 35.74 & 47.04 & 43.30 \\
\midrule
TPT~\citep{TPT} & 60.74 & 26.67 & 54.70 & 59.11 & 35.09 & 47.26 & 43.89 \\
C-TPT~\citep{C-TPT} & - & 25.60 & 54.80 & 59.70& 35.70 & - & 43.95  \\
DiffTPT~\citep{DiffTPT}  & 60.80 & 31.06 & 55.80 & 58.80 & {37.10} & 48.71 & 45.69 \\
TDA~\citep{TDA} & 61.35 & 30.29 & 55.54 & \underline{62.58} & \underline{38.12} & 49.58 & 46.63 \\
DPE~\citep{DPE} & 63.41 & 30.15 & 56.72 & \textbf{63.72} &  \textbf{40.03} & 50.81 & \underline{47.66} \\
TPT~\citep{TPT}+CoOp & \underline{64.73} & 30.32 & 57.83 & 58.99 & 35.86 & 49.55 & 45.75 \\
DiffTPT~\citep{DiffTPT}+CoOp & 64.70 & \underline{32.96} & \textbf{61.70} & 58.20 & 36.80 & \underline{50.87} & {47.42} \\
\midrule
\textbf{FGA (Ours)} & \textbf{65.90} & \textbf{37.87} &  \underline{59.03} & {62.32} & {37.63} & \textbf{52.55} & \textbf{49.21} \\
\bottomrule
\end{tabular}}
\label{tab:DG_RN50}
\end{table*}

\setlength{\tabcolsep}{1mm}
\begin{table*}[t]
\small
\centering
\caption{\textbf{Cross-dataset generalization with CLIP-ResNet50.} The models are fine-tuned on ImageNet with 16-shot training data per category. 
We emphasize the best results in \textbf{bold} and mark the second-best results with \underline{underlining}.
}\label{tab:2}
\scalebox{0.8}{\begin{tabular}{l|ccccccccc|c}
\toprule
Method & Caltech101 & Pets & Cars & Flowers102 & Aircraft & SUN397 & DTD & Food101 & UCF101 & Avg. \\
\midrule
CLIP-ResNet50~\citep{CLIP} & 87.26 & 82.97 & 55.89 & 62.77 & 16.11 & 60.85 & 40.37 &  74.82 & 59.48 & 60.06\\
\midrule
CoOp~\citep{coop} &  86.53 & 87.00 & 55.32 & 61.55 & 15.12 & 58.15 & 37.29 & 75.59 & 59.05 & 59.51 \\
CoCoOp~\citep{cocoop} &  \underline{87.38} & 88.39 & 56.22 & 65.57 & 14.61 & 59.61 & 38.53 & 76.20 & 57.10 & 60.40\\
\midrule
TPT~\citep{TPT} & 87.02 & 84.49 & 58.46 & 62.69 & 17.58 & 61.46 & \underline{40.84} & 74.88 & \underline{60.82} & 60.92\\
DiffTPT~\citep{DiffTPT} & 86.89 & 83.40 & \textbf{60.71} & \underline{63.53} & \underline{17.60} & \underline{62.67} & 40.72 & \textbf{79.21} & \textbf{62.27} & \underline{61.89} \\
TPT~\citep{TPT}+CoOp & 86.90 & \underline{87.54} & 57.65 & 58.83 & 15.84 & 60.00 & 38.06 & \underline{76.21} & 60.72 & 60.19 \\
\midrule 
\textbf{FGA (Ours) }& \textbf{89.74} & \textbf{88.50} & \underline{59.98} & \textbf{63.78} & \textbf{19.74} & \textbf{63.21} & \textbf{41.97} &  {75.43} & {60.66} & \textbf{62.56}\\
\bottomrule
\end{tabular}}
\label{tab:cross-dataset-RN50}
\end{table*}

\section{Reproducibility}\label{secD}
To guarantee reproducibility, we will provide an explanation of the code and hyperparameters in this section. 
\subsection{Code}\label{section:5}

Our work builds upon two key methodologies in prompt learning for vision-language models: CoOp (Conditional Prompt Learning)~\footnote{\url{https://github.com/KaiyangZhou/CoOp}} and TPT (Test-Time Prompt Tuning)~\footnote{\url{https://github.com/azshue/TPT}}, both of which are open-source projects released under the permissive MIT license. CoOp provides a foundation for adapting pre-trained models like CLIP to downstream tasks through learnable prompts, while TPT extends this by enabling dynamic prompt optimization during inference to enhance zero-shot generalization. All experiments in this study were performed on a single NVIDIA Tesla V100 GPU.

\subsection{Hyperparameters}\label{section:6}

\textbf{For $\boldsymbol{R}$:} 
We explore the impact of \(R\) on test accuracy with \(R \in \{2,4,8,16,32,64\}\). The results show minimal sensitivity (fluctuations $<$0.5\%) to \(R\) for \(R \geq 4\).
This insensitivity arises because we only utilize the relative value of sharpness for sample selection, which alleviates the need for precise calculations.
We finally set $R=16$ for all experiments.

\textbf{For $\boldsymbol{s}$:} 
We set $s$ to retain only 10\% of all augmented versions, aligning with the approach used in TPT, where 10\% of augmented versions are utilized for prompt optimization. It rules out any potential effects introduced by using more augmented data and ensures a fair comparison.

\section{Utilization of LLMs}\label{secE}
In academic writing, Large Language Models (LLMs) are used for linguistic refinement, such as correcting grammatical errors, improving word choice, and polishing sentence structures to enhance clarity and fluency. They are not involved in core scientific research activities (including idea generation, data interpretation, or substantive analysis), ensuring that all key contributions remain entirely human-generated.

\end{document}